\documentclass[pdflatex,sn-basic]{sn-jnl}% Basic Springer Nature Reference Style/Chemistry Reference Style

\usepackage{graphicx}%
\usepackage{multirow}%
\usepackage{amsmath,amssymb,amsfonts}%
\usepackage{amsthm}%
\usepackage{mathrsfs}%
\usepackage[title]{appendix}%
\usepackage{xcolor}%
\usepackage{textcomp}%
\usepackage{manyfoot}%
\usepackage{booktabs}%
\usepackage{algorithm}%
\usepackage{algorithmicx}%
\usepackage{algpseudocode}%
\usepackage{listings}%
\usepackage{tcolorbox}
\usepackage{ulem}
\usepackage{subcaption}
\usepackage{soul}

\begin{document}

\title[EMBRACE: Evaluation and Modifications for Boosting RACE]{EMBRACE: Evaluation and Modifications for Boosting RACE}

\author*[1]{\fnm{Mariia} \sur{Zyrianova}}\email{mariiaz@kth.se}

\author[1]{\fnm{Dmytro} \sur{Kalpakchi}}\email{dmytroka@kth.se}

\author[1]{\fnm{Johan} \sur{Boye}}\email{jboye@kth.se}

\affil*[1]{\orgdiv{Division of Speech, Music and Hearing}, \orgname{KTH Royal Institute of Technology}, \orgaddress{\street{Lindstedtsvägen 24}, \city{Stockholm}, \postcode{10044}, \country{Sweden}}}

\abstract{When training and evaluating machine reading comprehension models, it is very important to work with high-quality datasets that are also representative of  real-world reading comprehension tasks. This requirement includes, for instance, having questions that are based on texts of different genres and require generating inferences or reflecting on the reading material.

In this article we turn our attention to RACE, a dataset of English texts and corresponding multiple-choice questions (MCQs). Each MCQ consists of a question and four alternatives (of which one is the correct answer). RACE was constructed by Chinese teachers of English for human reading comprehension and is widely used as training material for machine reading comprehension models. By construction, RACE should satisfy the aforementioned quality requirements and the purpose of this article is to check whether they are indeed satisfied.

We provide a detailed analysis of the test set of RACE for high-school students (1045 texts and 3498 corresponding MCQs) including (1) an evaluation of the difficulty of each MCQ and (2) annotations for the relevant pieces of the texts (called \textit{bases}) that are used to justify the plausibility of each alternative. A considerable number of MCQs appear not to fulfill basic requirements for this type of reading comprehension tasks, so we additionally identify the high-quality subset of the evaluated RACE corpus. We also demonstrate that the distribution of the positions of the bases for the alternatives is biased towards certain parts of texts, which is not necessarily desirable when evaluating MCQ answering and generation models.}


\maketitle

\section{Introduction}\label{sec:intro}
A common way of assessing reading comprehension is to have students read a text and then let them answer questions on the contents. When answering the questions, the students have to combine facts they have obtained from the text with common-sense reasoning and, possibly, some prior knowledge. This is not only a challenging task for students learning a language, but is also a touchstone of machine natural-language understanding. A computer program that can process (``read'') a text and then accurately answer questions about it must be considered to have attained some level of language understanding. Hence, the interest in this task in the NLP community has grown considerably recently, which manifests itself in a number of published datasets used for training and evaluating models for machine comprehension of text (e.g., \cite{squad,coqa,quac}). 

When addressing the above-mentioned NLP tasks, it is of paramount importance that the datasets used for training and evaluation are not just large but also varied in genre, with questions of varying levels of difficulty, reflecting the structure of the real-world human tests of reading comprehension. Human reading comprehension tasks are constructed by teachers and other experts and tend to target more advanced reading processes, such as integrating information from different parts of the text, generating inferences, or reflecting on the reading material \citep{oecd2019pisa}.  To properly test the reading comprehension abilities of machines, they should be tried on such challenging questions. However, a dataset like SQuAD, which is constructed only to act as training material for automatic reading comprehension, uses texts only of one genre, e.g., texts from Wikipedia. On top of that, the questions in such datasets tend to be purely factual, and are typically constructed by crowd workers, instead of teachers.

Taking the aforementioned reasons into account, researchers should turn their attention to datasets with tasks resembling those for human reading comprehension. One such dataset for English is RACE \citep{lai-etal-2017-race}. This dataset is based on a large number of multiple-choice questions for reading comprehension, collected from \emph{real-world} English examinations for Chinese middle-school and high-school students. RACE is widely used by the community, for instance, as a training material for {\it automatic reading comprehension} models by numerous researchers (e.g., \cite{Xu2017DynamicFN}, \cite{Wang2018ACM}, \cite{Ran2019OptionCN}, \cite{zhang2020dcmn+}, \cite{Lin2021UsingAA}). Researchers have also used RACE to address the task of {\it question generation} by training models that attempt to automatically generate reading comprehension questions from texts (e.g., \cite{Jia2020EQGRACEEQ}, \cite{steuer2020}, \cite{Lelkes2021QuizStyleQG}). Similarly, RACE has been used to generate whole multiple-choice questions from the text, i.e. generate the question and alternatives, given a text passage (and possibly a correct answer) (e.g., \cite{Gao2018GeneratingDF}, \cite{Zhou2019CoAttentionHN}, \cite{Offerijns2020BetterDT}, \cite{qiu-etal-2020-automatic}, \cite{chung-etal-2020-bert}). Additionally, a new dataset, RACE++ \citep{liang2019new}, has been built on top of RACE, and models have also been trained on this new dataset, e.g. \cite{raina2022multiplechoice,khashabi2022unifiedqa,liusie2022world,rosati2022using}. 

%Since we are interested in \emph{human} reading comprehension (later referred to as simply \textit{reading comprehension or RC}), 
In this article, we turn our attention to RACE \citep{lai-etal-2017-race}, not least because of its wide adoption by the research community. More specifically, we evaluate the quality of a subset of RACE, since it is of paramount importance that the datasets used for training and, especially, evaluation are not just large but also of high quality for any of the aforementioned NLP tasks. Before diving into what ``high quality'' entails, recall that the tasks in RACE are in multiple-choice format. This means that each task contains a question (the {\it stem\/}) based on a given text, and a number of alternatives, of which one is the correct answer (called the {\it key\/}), and the others are incorrect but (hopefully) plausible alternatives (called the {\it distractors\/}). For this format, basic quality requirements include that: (1) the text, stem, and all alternatives must be grammatically correct; (2) the key should indeed constitute the correct answer for the stem; (3) the stem should {\it not\/} be answerable without reading the text (i.e. the text should be both {\it sufficient\/} and {\it necessary\/}), and (4) \textit{all} distractors for each question should indeed be wrong. On top of that, professional test constructors impose additional quality requirements. For instance, \cite{haladyna2004developing} specifies that distractors should be distinct and non-overlapping, and all alternatives should have the same grammatical structure and approximately the same length in order not to give meta-clues as to which alternative is correct. Using all aforementioned criteria as guidelines, we analyse the test set of RACE for high-school students and point out that it has a number of text passages and questions that do not fulfill these crucial requirements. One of our endeavors is to {\it identify the high-quality subset of this test set\/} (see Section \ref{sec:varerr}) for evaluating models for both automatic reading comprehension and automatic question generation tasks.

Furthermore, some reading comprehension questions are easier to answer than others, and this is the case both for human test-takers and for models of human reading comprehension. It is important to take this difficulty into consideration when evaluating a reading comprehension model. Perhaps the model manages to answer many questions correctly but, on closer scrutiny, it is revealed that the model only manages to answer easy questions and fails on more challenging ones. In this article, we use and adapt difficulty model originally developed for reading comprehension questions by \cite{kirsch1995interpreting}. Using this model, we extend the analysed subset of RACE with additional annotations that mark difficulty of the included multiple-choice questions (see Section \ref{sec:Koverv}). On top of that, we also mark the pieces of text which allow grounding the alternatives in the text (see Section \ref{sec:addan}).

The contributions of this article concern the test set of RACE for high-school students (later referred to as the \textit{RACE corpus}) and, more specifically, include:
\begin{itemize}
    \item a detailed analysis of the quality of texts (Sections \ref{sec:vartext}, and \ref{sec:varerr}) and MCQs (Sections \ref{sec:varmcq} and \ref{sec:varerr});
    \item additional annotations for MCQ difficulty (Section \ref{sec:mcqdiff}) based on the evaluation scheme originally developed by \cite{kirsch1995interpreting,kirsch1999,kirsch2001international} and adapted as explained in section \ref{sec:Koverv};
    \item additional annotations for pieces of texts that justify the plausibility of each alternative in the text, called \textit{bases} (Section \ref{sec:addan}).
\end{itemize}
Both annotations and the source code to reproduce the analysis are available online at \url{https://github.com/dkalpakchi/EMBRACE}.

\section{Overview of the RACE dataset}\label{sec:race}
The \textbf{ReAding Comprehension dataset from Examinations (RACE)} is a large-scale reading comprehension (RC) dataset collected of multiple-choice questions (MCQs) from English examinations in China. The corpus is divided into two parts: RACE-M, containing text passages and questions designed for 12–15 year-old (middle school) students, and RACE-H, containing questions and text passages for and 15–18 year-old (high school) students. The dataset includes 27933 passages (7139 in RACE-M and 20794 in RACE-H), and 97687 MCQs (28293 in RACE-M and 69394 in RACE-H) to test EFL (English as a Foreign Language) learners' reading comprehension skills. 

\textbf{Collection of the MCQs:} The data were collected from examinations published on free public websites in China. Additionally, \citet{lai-etal-2017-race} filtered the collected data by removing: (1) all articles and questions that are not self-contained based on the text information (those containing images or tables); (2) all duplicate articles; (3)~MCQs where the number of answer alternatives was not four, and (4) MCQs containing the keywords “underlined” or “paragraph”, since ``it is difficult to reproduce the effect of underlines and the paragraph segment information'' \citep{lai-etal-2017-race}.

\textbf{Properties of the MCQs:}  All questions and alternatives were written by human experts. Each question is either in the form of an interrogative sentence or as a fill-in-the-gap task, and is provided with \textit{four answer alternatives}, of which \textit{only one is correct}. Furthermore, the alternatives are ``human generated sentences which may not appear in the original passage'' (\cite{lai-etal-2017-race}). 

\textbf{Classification of the MCQs:} Following \cite{chen-etal-2016-thorough} and \cite{trischler-etal-2017-newsqa}, \cite{lai-etal-2017-race} classified the ``reasoning/question types'' into five groups, with ascending order of difficulty: \textit{word matching, paraphrasing, single-sentence reasoning, multi-sentence reasoning, insufficient/ambiguous}. The proportion of different reasoning/question types was then estimated by having humans label 500 passages (250 from RACE-M and RACE-H each). The most frequent subdivisions of questions under the ``reasoning'' categories included \textit{detail reasoning, whole-picture understanding, passage summarization, attitude analysis, and world knowledge}, with a possibility for an MCQ to fall into multiple categories.

Both RACE-M and RACE-H were split 90/5/5  into training, development, and test sets respectively. In this paper, we only consider the \textbf{\textit{test set for RACE-H}} (which we will later refer to as the \underline{RACE corpus}), encompassing
\textit{1045 passages and 3498 questions}.

\section{Terminology}
Before we dive into details of the RACE evaluation we establish the following terminology to be used throughout the rest of the article:
\begin{itemize}
    \item\textbf{MCQ} refers to the combination of the \textit{text and an MCQ unit} based on this text. This is to emphasise that a text is the integral part of any reading comprehension task and MCQ units that satisfy the quality requirements are meaningless in absence of the corresponding text.
    \item \textbf{MCQ unit} refers to the combination of the \textit{question}\footnote{the same as ``stem'', the terms are used interchangeably in this article}, the \textit{correct answer}\footnote{the same as ``key'', the terms are used interchangeably in this article} to this question, and incorrect but plausible options called \textit{distractors}.
    \item \textbf{Alternatives} refer to the the \textit{correct answer and distractors} together.
    \item \textbf{MCQ element} refers separately to the \textit{text, question, or alternatives}, all given for one MCQ.  
\end{itemize}

\section{Overview of the evaluation model}\label{sec:Koverv}
To assess the complexity of the given texts and corresponding MCQ units, we mainly adopt an evaluation model and scoring rules developed by  \cite{kirsch1995interpreting,kirsch1999,kirsch2001international}. 
The evaluation model decomposes an MCQ along different dimensions and assigns a numerical difficulty score for each MCQ per dimension. The total difficulty score for an MCQ can be obtained by summing up the per-dimension difficulties for this MCQ. 

The Kirsch-Mosenthal scheme was initially developed for testing the reading comprehension skills of native English-speaking students. We are modifying and extending the scheme slightly since the texts are more diverse in RACE and RACE is also aimed at second language learners of English. We present our interpretation of the scheme in section \ref{sec:varcond}. 

In order to evaluate the MCQs from the RACE-H test set, we first identify the characteristics of the text (\ref{sec:divtext}) and then assess each MCQ according to the variables included in the evaluation model (\ref{sec:varcond}).

\subsection{Diversity of texts} \label{sec:divtext}
The texts appearing in the RACE corpus were collected from English exams for middle-school and high-school Chinese students within the 12–18 age range. \cite{lai-etal-2017-race} state that ``passages in RACE almost cover all types of human articles, such as news, stories, ads, biography, philosophy, etc., in a variety of styles''. 
Following \cite{kirsch2001international,oecd2019pisa}, we classify the texts in RACE as belonging to one of the categories below:
\begin{itemize}
    \item \textbf{\textit{continuous texts}} are texts composed of sentences organised into paragraphs, with no lists, tables, or graphs, and without subheadings to paragraphs or other sequence markers to indicate relations between units of the text. Types of texts treated as ``continuous'' are \textit{narration, description, exposition, argumentation}.\\
    
    \item \textbf{\textit{partly continuous texts}} are texts containing headers for paragraphs, or texts that include numbered lists, bullet-point lists or similar, in addition to ordinary continuous paragraphs.  Partly continuous texts include \textit{instructions, documents, advertisements, catalog descriptions}.\\
    \item \textbf{\textit{non-continuous texts}} are so-called ``matrix documents'' \citep{kirsch2001international}, which are formatted as a list, table, schedule, chart, graph, map, or form.\\
    \item \textbf{\textit{mixed texts}} would include a paragraph together with a picture, or a graph with an explanatory map, so that elements of both continuous and non-continuous formats are present in one passage. However, the developers of the RACE dataset filtered out all texts and questions containing images or tables (texts ``not self-contained based on the text information''). 
\end{itemize}
For the distribution of the texts in the RACE corpus, see Section \ref{sec:vartext}, Figure \ref{fig:textformat}.

The broad genre and style coverage of the RACE corpus is useful when evaluating the reading comprehension abilities of computer systems, but as pointed out by \cite{kirsch1995interpreting,kirsch1999}, \cite{oecd2019pisa}, different text styles call for different ways of assessing the difficulty of texts and MCQ units based on the texts. For the evaluation purposes of this research, we primarily focus on continuous texts, while also providing annotations for partly continuous texts to the best of the ability of the applied evaluation model. We exclude non-continuous texts from the evaluation process as those requiring different reading and evaluation approaches for a number of dimensions \citep{oecd2019pisa,kirsch2001international}.

\subsection{Evaluation variables} \label{sec:varcond}
Below we explain the nature of each variable and provide some examples of how one is represented in the RACE corpus, and state the scoring rules we apply for the assessment. These variables are: \textit{Type of information, Type of match, Number of phrases, Number of items, Number of items transparency, Number of required paragraphs, Infer condition, Plausibility of distractors, Type of calculation}. For more detailed explanations, specific cases, and examples see \cite[Section 2]{zyrianovam}.

\subsubsection{Type of information (TOI)} \label{sec:toi}
%from Word draft 
TOI refers to the kinds of information that readers need to identify in order to answer a question successfully. The more abstract the requested information is, the harder the task is thought to be. Hence, a numerical score is awarded depending on how abstract the concept is, ranging from ``1'' for the very concrete to ``5'' for the very abstract. Table~\ref{tab:toi} presents all the original categories and their scores used in at least one of the articles \citep{kirsch1995interpreting,kirsch2001international,kirsch1999}.

\begin{table}[h!]
\centering
    
    \begin{tabular}{ |p{9.5cm}|p{2.5cm}|  }
     \hline
     \textbf{Concepts} & \textbf{Points/Category} \\
     \hline
     person, animal, place, group, thing & 1 \\
     \hline
     amount, time, attribute, action, location, type/kind, procedure, part & 2 \\
     \hline
     manner, goal, purpose, condition, predicate adjective, function, alternative, attempt, sequence, pronominal reference, verification, assertion, problem, solution, role, process  & 3 \\
     \hline
     cause, reason, result, effect, justification, evidence, similarity, opinion, explanation, theme, pattern & 4 \\
     \hline
     equivalent, difference, definition, advantage; indeterminate & 5 \\
     \hline
   
\end{tabular}
\caption{Categorisation of the TOI concepts applied to the RACE corpus evaluation}
    \label{tab:toi}
\end{table}

However, we also identify a number of concepts used as type-of-information in RACE, but which are not present in Table~\ref{tab:toi}, so we add these to the scheme. While doing so, we follow the original principle: the more concrete the concept, the easier it is (and hence the lower TOI score it gets). The final version of the scheme includes 5 categories with altogether 42 different concepts (4+8+16+11+5). 

\begin{table}[h!]
\centering
    
    \begin{tabular}{ |p{5.5cm}|p{4cm}|p{2.5cm}| }
     \hline
     \textbf{New TOI concepts} & \textbf{Mapped to} & \textbf{Points/Category} \\
     \hline
     proper names (people) & person & 1 \\
     \hline
     proper names (published materials, names of events) & thing & 1 \\
     \hline
     contact details (telephone number, email address) & thing& 1 \\
     \hline
     job, profession, position & person / group / role 
– \textbf{context-dependent}
 & 1 / 1 / 3 \\
     \hline
     age & amount & 2 \\
      \hline
     prerequisite & condition & 3 \\
      \hline
     attitude & opinion & 4 \\
      \hline
     recommendation, piece of advice & opinion & 4 \\
      \hline
     main idea, purpose of the passage & theme & 4 \\
      \hline
     example & equivalent & 5 \\ 
     \hline
   
\end{tabular}
\caption{TOI concepts that were added for the RACE corpus evaluation}
    \label{tab:toiadj}
\end{table}

Identification and categorization of the type of information are mainly straightforward, since in most cases it can be derived from a combination of the question word (e.g., \textit{“When?” – time; “How many?” – amount}, etc.) and the answer alternatives. For examples of tricky cases, see \cite[Section 2.1]{zyrianovam}. 

In some MCQs in RACE, the answer alternatives belong to different types. In cases where the alternatives belong to 4 different types, or where 2 alternatives belong to one type and the 2 other alternatives belong to another, we consider the TOI to be "indeterminate" and award a TOI score of 5. Cases where 3 alternatives belong to one type and the remaining alternative belongs to another were considered to be errors (labeled as X in Figure \ref{fig:choicesproblems}) and excluded from further analysis. The reason for this is that such a combination of alternatives might create a bias, and allow the student to detect the correct answer without reading the text, which is highly undesirable for a reading comprehension test (see more in \ref{sec:varerr}).

\textbf{Note:} for TOI, grammatical number (\textit{singular – plural}) of items of the same domain (e.g., \textit{place – places, function – functions, similarity – similarities} etc.) has \underline{no} impact on the evaluation and is controlled by another variable (see \ref{sec:ni}).

See complete scale applied for the RACE corpus evaluation in Appendix \ref{app:A}.

\subsubsection{Type of match (TOM)} \label{sec:tom}
TOM deals with the degree of difficulty associated with matching information in a question to information in a passage \citep{kirsch1995interpreting}. \cite{kirsch1995interpreting,kirsch1999}). According to the rules, the relations between the text and the question (further referred to as \textit{T-Q}) and the text and the answer (further referred to as \textit{T-A}) are viewed in terms of the degree of lexical and grammatical correspondence and are, therefore, represented by one of the following types: \textit{literal match}. \textit{synonymous match}, \textit{low-level text-based inference}, \textit{high-level text-based inference}, or \textit{generating the appropriate interpretive framework}.

\cite{kirsch1995interpreting} provided no definitions for mentioned types of T-Q and T-A relations, but illustrated the classification with examples of RC texts and corresponding tasks for 4th and 9th grade schoolchildren, including final scores for each of the questions. This allowed us to relate the represented questions with their categorisation and, basing on that, define TOM categories for further evaluation of the RACE corpus. The classification we follow for the purposes of the RACE corpus evaluation is as follows, in increasing level of difficulty:
\begin{itemize}
\item \textbf{Literal match (LM)} means that a piece of information given in the text corresponds word-by-word to one of the alternatives (in the case of T-A), or in the question (in the case of T-Q). In addition to \textit{verbatim correspondence}, special cases include, among others, 
\textit{contractions}, \textit{incomplete correspondence between proper names} (e.g.\ a person is mentioned with full name in the text, but only with the family name in the question), and \textit{paronymous adjectives with close meaning} (e.g., historic -- historical). 
See \cite[Section 2.2.1]{zyrianovam} for the complete list with examples.

\item \textbf{Synonymous match (SM)} is a type of T-Q or T-A relation where one or more word(s) or phrase(s) are substituted by distinct word(s) or phrase(s) respectively, with the meaning kept identical to that in the text. The main types of SM are \textit{vocabulary replacements}, \textit{grammatical alterations}, and \textit{cases when numbers are written in words in the text, but with digits in the question or alternatives (or vice versa)}.
%and/or a word family involvement; numbers written in words or digits, and used interchangeably. 
See \cite[Section 2.2.2]{zyrianovam} for examples.

\item \textbf{Low-level text-based inference (LLTI)} refers to such type of T-Q or T-A relations when it is impossible to find a literally or synonymously matching piece of information in the text. Instead, the requested information should be integrated from one or more phrases, sentences or paragraphs. For RACE evaluation, integration was defined by \textit{resolving pronominal reference(s), drawing inference(s) within several clauses/sentences, inferring the definition given a term (or vice versa), recognising patterns, applying cause-and-effect reasoning, making transformations from/to the opposite, comparing/contrasting two or more items from the text}, or \textit{finding contrast within alternatives}. Note that the aforementioned categories are not mutually exclusive, and if MCQ exhibits multiple of those, it is not scored higher than LLTI. See \cite[Section 2.2.3]{zyrianovam} for explanations and examples.

\item \textbf{High-level text-based inference (HLTI)} refers to the type of T-Q or T-A relations that either \textit{require skills other than reading} (e.g., to perform calculations), or necessitate \textit{a specific type of background knowledge} (e.g., to connect ethnonyms to the geographical locations, e.g., Dutch -- the Netherlands, or to understand mathematical terminology), or test the reader's ability to apply or distill \textit{knowledge gained from reading} (e.g., to solve a hypothetical problem not mentioned in the passage, or to understand the purpose of the author). In contrast to LLTI, the text provides no explicit clues as to which pieces of it are particularly relevant for the MCQ. Instead, readers must determine the relevant parts themselves. See \cite[Section 2.2.4]{zyrianovam} for explanations and examples.

\item \textbf{Generation of the appropriate interpretive framework (GEN)} implies that high-level text-based inference is needed \textit{both} for matching the text and the question (T-Q relations) and for matching the text and the answer (T-A relations). 
See \cite[Section 2.2.5]{zyrianovam} for examples. 
\end{itemize}
Scoring rules for the TOM variable were specifically developed for continuous texts by \cite{kirsch1995interpreting} (a different scoring model was devised for partly continuous texts by the same authors). The more matching, inferencing, or integration of information is required, the higher the TOM score is assigned to the MCQ.
For the RACE corpus evaluation, we used the scale summarised in Table \ref{tab:tom} with its detailed version presented in Appendix \ref{app:A}. Note that if any relations from Table \ref{tab:tom} are swapped (e.g., the T-Q relation is ``synonymous match'' and the T-A relation is ``literal match''), the number of awarded points will remain the same (1 point, in this case).

\begin{table}[h!]
\centering
    \begin{tabular}{ |p{4.5cm}|p{4.5cm}|p{3cm}|  }
     \hline
     \textbf{T-Q relations} & \textbf{T-A relations} & \textbf{Points} \\
     \hline
     literal  match & literal match & 0.5 \\
      \hline
     literal  match & synonymous match & 1 \\
     \hline
     synonymous match & synonymous match & 1.5 \\
      \hline
     literal match & low-level text-based inference & 2 \\
      \hline
     synonymous match & low-level text-based inference & 2.5 \\
      \hline
     low-level text-based inference & low-level text-based inference & 3 \\
     \hline
     LM or SM or LLTI & high-level text-based inference & 4 \\
     \hline
     \multicolumn{2}{|c|}{generate the appropriate interpretive framework} & 5 \\
     \hline 
    \end{tabular}
    \caption{The summarised TOM scoring scale applied for the RACE corpus evaluation}
    \label{tab:tom}
\end{table}

\subsubsection{Number of phrases (NPhr)} \label{sec:nphr}
NPhr, \textit{the number of} included \textit{independent and dependent clauses} in the question, is important to take into account since each such clause could bring an additional condition/trait to control when determining the correct answer (\cite{kirsch1999}). Insofar as \textit{detached predicatives, parentheticals, prefaces} \citep{Biber_99} state additional criteria to be satisfied in the correct answer, each of the three is counted as a phrase for NPhr. 
See \cite[Section 2.3]{zyrianovam} for examples.

On this variable, MCQs score from 0 to 3 points (the less phrases there are, the lower the score is). See complete scale applied for the RACE corpus evaluation in Appendix \ref{app:A}.

\subsubsection{Number of items (NI)} \label{ni}\label{sec:ni}
NI is the number of parts included to the correct, such that each part separately could serve as a relevant (but not complete) answer to the question. For instance, in case "Bob and Jane" is the correct answer, NI is 2.

For MCQ units where the correct answer consists of a combination of other two or three alternatives given to the same MCQ (e.g., \textit{both A and B, both A and C, all of the above}, etc.), which are referred to as \textbf{\textit{complex alternatives}} in this paper, NI is counted according to the number of items included in the correct answer. See \cite[Section 2.4]{zyrianovam} for more details and examples. 

Depending on the NI in the correct answer, MCQs score from 0 to 3 points (see Appendix \ref{app:A} for the complete scale applied for the RACE corpus evaluation).

\subsubsection{Number of items transparency (NIt)} \label{nit}
The variable indicates if NI in the correct answer is specified for a reader or can be derived from the MCQ unit structure. It is \textbf{\textit{specified}} if either the number of items is given in the question (e.g., ``{\it Which \underline{two} people...?}''), or if all the alternatives have the same number of items, see \cite[Section 2.5]{zyrianovam} for more examples.
In both these cases, the NIt is awarded 0 points. In other cases, NIt is assessed as \textbf{\textit{unspecified}} and the MCQ is awarded 1 point.

\subsubsection{Number of required paragraphs (NPar)} \label{sec:np}
The variable represents the \textit{minimal} number of paragraphs which is enough for a reader to make one or more corresponding match(es) or inference(s) between the text and question or the text and correct answer. For the RACE evaluation, both paragraphs required to relate the text to the question, and paragraphs required to relate the text to the correct answer are counted. Inferences made within one paragraph are considered easier than those requiring the reader to combine information from two or more paragraphs. Hence, MCQs of the first-mentioned case score 0 points, while MCQs of the second-mentioned case receive 1 point.
See the complete scale applied for the RACE corpus evaluation in Appendix \ref{app:A}.

\subsubsection{Infer condition (IC)} \label{sec:ic}
Two types of relations between the question and the text devised by \cite{kirsch1995interpreting,kirsch1999} as possible infer conditions are \textit{compare} (search for similarities) and \textit{contrast} (search for differences), where ``contrast'' is considered more difficult than ``compare''. An example of ´´contrast'' would be when certain symptoms of a disease are described in the text, and the question asks to select what \underline{cannot} be viewed as a symptom of this illness. Questions for contrast often include negations like \textit{not, except}, or prefixes with negative connotation like \textit{dis-, un-, im-, ir-} etc.  The compare/contrast distinction was introduced as an additional condition (IC) to be assessed \citep{kirsch2001international}, and we have kept is as-is for the RACE corpus evaluation.

The IC score is either 0 (for easier cases) or 1. The scoring of IC is influenced by the NPar variable, and the exact scoring rules are described in detail in Appendix \ref{app:A}.

\subsubsection{Plausibility of distractors (POD)} \label{sec:pod}
POD represents the degree of difficulty associated with selecting the correct answer from the list of alternatives \citep{kirsch1995interpreting}.
The principle of evaluation of the variable implies that the more conditions the distracting information shares with a correct answer and the closer it is positioned to the correct answer, the more difficult the processing becomes \citep{kirsch1999}. To assess this difficulty, the \textit{distance} between each of the distractors and the correct answer (if they both are in the same paragraph or not) is taken into account. In the case when all distractors appear in a different paragraph(s) from the correct answer, they are assessed then by \textit{type of match} – each by the same principles as T-A relations for TOM; if at least one of the distractors appears in the same paragraph with the correct answer,  the \textit{number of} such \textit{distractors} is taken for evaluation. 
See \cite[Section 2.8]{zyrianovam} for more details and examples.

The scoring principles mandate that the easiest distractors (the furthest from the correct answer in matching context and distance) score the least possible points, which is ``1'', while the most difficult distractors (the closest to the correct answer, or those representing plausible inferences based on information outside the text) gain the maximum, which is ``5''. For the options where literal or synonymous match is mentioned, they gain 2 and 2,5 points respectively. 
See complete evaluation scale applied for the RACE corpus evaluation in Appendix \ref{app:A}.

\textbf{Note:} in case if distractors for the same MCQ appear in different assessment categories, \emph{the one to score the highest} is chosen for evaluation, since this will be the closest to the correct answer, representing the maximum number of conditions/traits to be verified and thus, reflecting the required minimum of knowledge and skills from a reader to detect the correct answer, see \cite[Figure 20]{zyrianovam} for an example.

\subsubsection{Type of calculation (TOC)} \label{sec:toc}
TOC was initially elaborated by \cite{kirsch1999} as an additional variable for quantitative literacy evaluation, mainly necessary for comprehension of noncontinuous texts or documents. The variable represented types of arithmetic operation \textit{(addition, subtraction, multiplication, or division)} and whether that operation must be performed alone or in combination. 

However, one-by-one counting, which appears in approximately 20\% of all calculation-related MCQs of the evaluated RACE materials (questions like \textit{“How many characters are mentioned in this story?”(262), ``How many people does Mr. Brown see in the street one day? He sees \_ in all.''(3425)}) was not initially discussed among the types of operation. As a solution for the RACE corpus evaluation, if no other operation(s) required, such MCQs get the same scoring as for \textit{single addition} (which is treated as the easiest),  since this type of calculation itself is basic for all other mathematical operations.

The final scoring ranges from 0 (no operations required) to 5 (multiple operations required). See Appendix \ref{app:A} for the complete evaluation model.

\subsubsection*{Modifications}
This section summarises all modifications we made to the model introduced by \cite{kirsch1995interpreting} and added by \cite{kirsch2001international}, in order to adapt the model to the evaluation purposes of this research. Despite the main intention from our side to keep the evaluation model as close to the original one \citep{kirsch2001international} as possible, a need for adaptions arises from the differences in (1) target groups RC tests were composed for, (2) specificity of the included texts, or (3) specificity of the included MCQs.

Thus, the modifications include the following:
\begin{enumerate}
    \item \textit{For TOM and POD assessment}, the notion of \textbf{literal or synonymous match} is split into two separate categories (\textbf{``literal match''} and \textbf{``synonymous match''}), with a consequent partition of all options where ``literal or synonymous match'' composes a part. 

    As it follows from \cite{oecd2021pisa}, \textit{word recognition} and \textit{lexical search} (basically represented in the evaluation model by \textit{literal} and \textit{synonymous match}, respectively) characterise two different in complexity
    %, though following one another, 
    stages for foreign language learners. That given, we split the notion of \textit{literal or synonymous match} into two separate categories (namely, \textit{literal match} and \textit{synonymous match}), with a consequent partition of all options where ``literal or synonymous match'' composes a part. Hence, the number of optional categories for TOM becomes \textit{eight}, instead of five initially developed; and the number of categories for POD becomes seven, instead of the six initially proposed by \cite{kirsch1995interpreting}, \cite{kirsch2001international}. 
    
    Consequently, we also adjust the scoring scale so that ``literal match'' scores the least possible points as the easiest category of TOM, and the proportional correlation between other categories still corresponds to that proposed by \cite{kirsch1995interpreting}.\\

    \item \textit{Maximum} \textbf{number of phrases (NPhr)} mentioned in the corresponding options for the variable is \textbf{altered from \textit{four} (``4'') to  \textit{four or more} (``4+'')}.

    The original scoring model \citep{kirsch1995interpreting,kirsch1999,kirsch2001international} was developed to evaluate questions that include no more than four clauses. However, some questions in RACE do include more. As a solution for assessment purposes, we represent the NPhr scale similarly to that for NI (adapted from the original model), where the maximum also allows all higher numbers to score the same (i.e. five or more items in the correct answer result in 3 points for the MCQ). Hence, questions to include more than four phrases score 3 points each for the RACE corpus evaluation.\\

    \item \textbf{Type of calculation (TOC)} is introduced as a process variable for the assessment of all kinds of texts from the RACE corpus. This allows us to detect MCQs that require mathematical skills in addition to reading skills, and award such MCQs with extra points for complexity. However, since calculation is not supposed to be common in RC tasks, MCQs where calculation is not required, score 0 points on this variable. No other alterations are made to the original scoring scale.
\end{enumerate}

\subsection*{Assessment principles}
Below we present the main principles we follow when assessing the RACE corpus according to the above-mentioned variables:

\begin{itemize}
    \item The assessment of each MCQ aims to \textbf{reflect the minimum} threshold for skills/knowledge required from a reader to get the correct answer.
    \item For cases where target MCQ element(s) satisfy conditions of \textbf{several different categories under the same variable} (except TOI), the one to score the highest is kept for further evaluation (e.g., for TOM, when T-A relations combine both LM and SM, SM is kept for annotation). 
    \item For MCQs where \textbf{TOI} can be characterised by several categories, the one to score the least is selected for annotation. 
    \item For MCQs with \textbf{multiple ways} of getting the correct answer, where \underline{final score} is \underline{the same} in all ways, the one based on the closest correspondence between the text and the correct answer is selected for annotation. 
    \item For MCQs with \textbf{multiple ways} of getting the correct answer, where the \underline{final score differs} depending on the followed way, the one to score the least is selected for annotation. 
    \item Annotation is considered complete when the \textbf{MCQ is assessed and scored under each variable} included in the model.
    
\end{itemize}

\section{Results of the RACE corpus evaluation}
The evaluation was conducted manually, mostly by the article's first author, who was assigned as the main annotator due to her relevant academic background in philology and practical experience in teaching EFL. To complete the evaluation process, we annotated the RACE corpus iteratively. All three researchers meet regularly to review and discuss annotation issues, after which re-annotation took place where necessary. All annotations were completed using an instance of Textinator \citep{kalpakchi2022textinator} as the annotation tool.

\subsection{Data quality} \label{sec:datqual} 
\subsubsection{Variety of texts} \label{sec:vartext}
Aside from the text format variety (defined in section \ref{sec:divtext}), RACE includes cases with either single or multiple texts for the same MCQ unit. In \cite{oecd2019pisa}, the former cases are called \textit{single-source texts} (those presented to the reader in isolation from other texts, even if they do not explicitly have any source indication), and the latter ones are called \textit{multiple-source texts} (those with different texts having different authors, or being published at different times, or by having different titles or reference numbers).

The texts from the RACE corpus do not include text descriptors (like date of publication, authors, and often -- titles). Hence, it is impossible to define sources as such, and instead we count the \textit{number of texts that could be viewed as independent} (we will refer to them as \textit{member texts}). To avoid confusion with the terminology used in \cite{oecd2019pisa}, we introduce the notions of \textit{single-member texts} (instead of single-source texts) and \textit{multiple-member texts} (instead of multiple-source texts). Specifically in the RACE corpus, multiple-member texts might include or be composed by two or more:
\begin{itemize}
    \item excerpts from interviews/surveys to represent opinions of several people, formatted by the same rules, usually including (full) names, age, or location (not to be confused with citations used as part of argumentation) (e.g., customers' reviews);
    \item sections to accumulate data about separate objects of the same type (e.g., books, shops, waterfalls, etc.), formatted by the same rules, often divided by (sub)titles -- potentially written by different authors and combined into one passage;
    \item independent complete passages of the same genre (e.g., letters, advertisements, etc.).
\end{itemize}

All other passages which do not satisfy the above-mentioned conditions are treated as single-member texts.

Multiple-member texts might have an introductory paragraph, as well as (sub)headers to distinct the member texts. Additionally, each member text might contain elements of a non-continuous text (marked as partly continuous then), which are defined by the same principles as for single-member texts (see Section \ref{sec:divtext}).

\begin{figure}[t!]
	\centering
	\includegraphics[width=\linewidth]{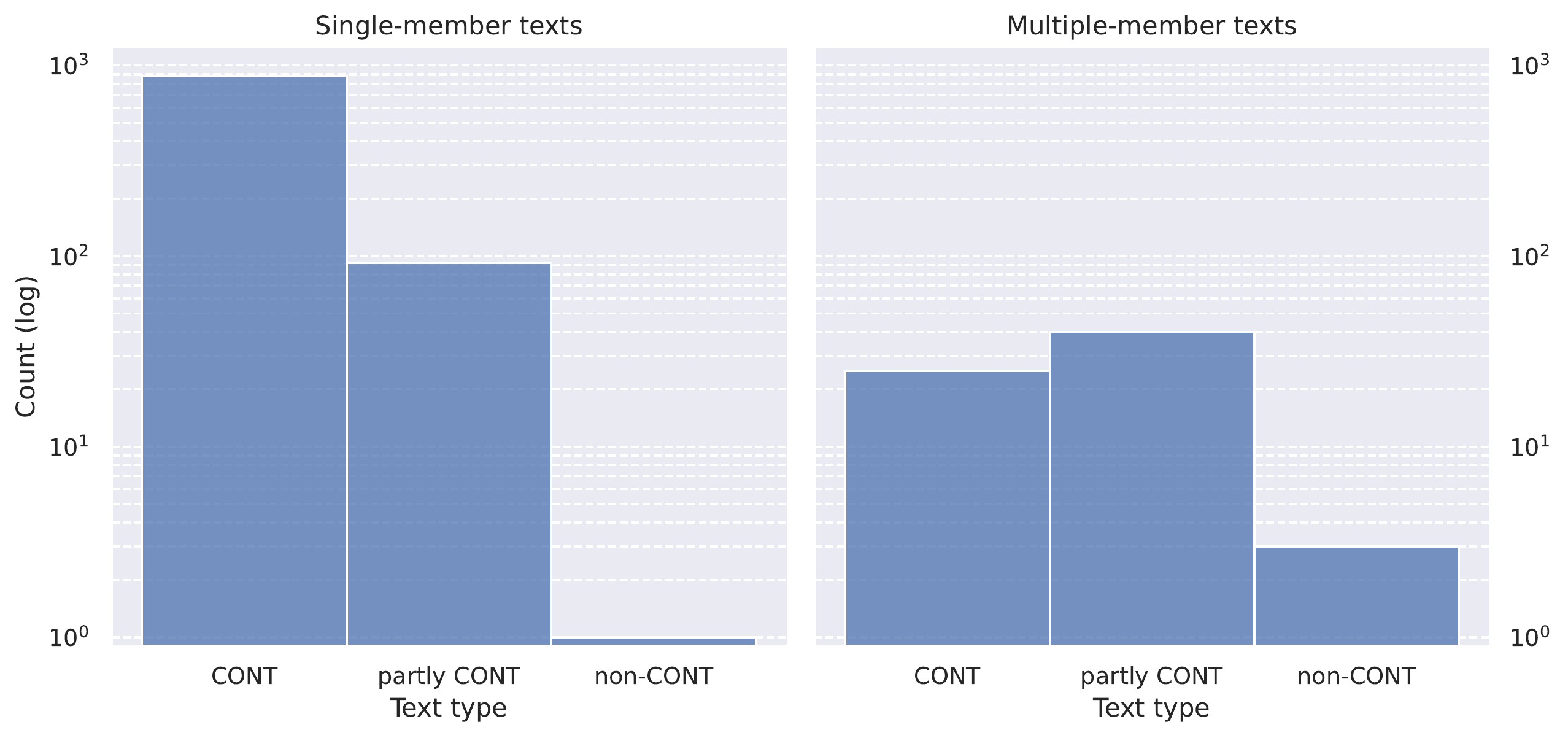}
	\caption{Distribution of text formats for the texts in the RACE corpus (in logarithmic scale). CONT stands for ``Continuous text''}
	\label{fig:textformat}
\end{figure}

The distribution of texts from the RACE corpus by the text format variety (as defined in section \ref{sec:divtext}) is presented in Figure \ref{fig:textformat}. We note that there are more single-member texts (the left part of Figure \ref{fig:textformat}) than multiple-member texts (the right part of Figure \ref{fig:textformat}). Overall, continuous texts dominate in the corpus by a substantial margin, although for multiple-member texts the number of partly continuous texts is the largest (albeit by a relatively small margin).

We also note that the total number of non-continuous texts in the whole corpus is much smaller (4 texts) compared to the other text types. These 4 passages are \emph{excluded} from the corpus (together with all corresponding MCQ units) because our evaluation scheme can not handle them. Hence, only 909 purely continuous texts + 132 partly continuous texts = 1041 texts are taken for further analysis.

\subsubsection{Variety of MCQ units} \label{sec:varmcq}
As mentioned in section \ref{sec:race}, multiple-choice tasks in RACE can be formulated as interrogative queries (e.g., \textit{What is the text mainly about?}) or fill-in-the-gap sentences (e.g., \textit{The text is mainly about \_ .}). In the RACE corpus taken for evaluation, we have dealt with both cases. 

In addition to this, the questions differ in the aspects of RC and language learning they focus on. Thus, the MCQ units can be based on (1) the content of the text, (2) text structure (titles and subtitles, division into paragraphs) or text descriptors (date of publication, author), or (3) the vocabulary used in a particular passage.  

\begin{figure}[!t]
	\centering
	\includegraphics[width=\linewidth]{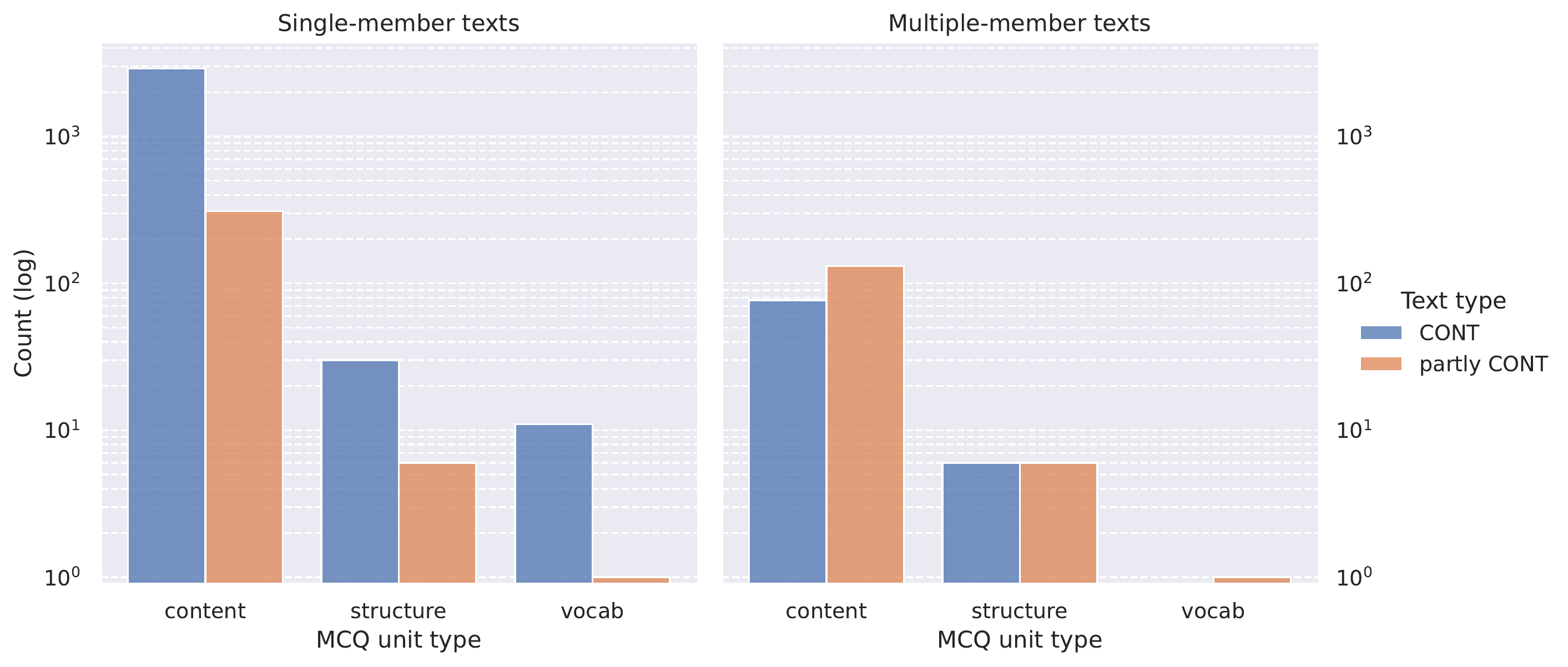}
	\caption{Distribution of the MCQ unit types (in logarithmic scale) depending on the type of texts. CONT = Continuous}
	\label{fig:mcqformat}
\end{figure}

The distribution of MCQ units by the aspects mentioned above is provided in Figure \ref{fig:mcqformat}. We note that content-based MCQ units dominate the corpus. Out of the total 3498 MCQ units, there are 3424 content-based ones, which correspond to 1038 of 1041 passages. The majority of such MCQ units (3216) are based on single-text materials, whereas only 208 MCQ units are based on multiple-text ones. There are substantially fewer structure-based MCQ units and even fewer vocabulary-based ones. 

These last two categories are \emph{excluded} from all further analysis, since (1) text descriptors are often excluded from the texts in RACE; (2) the applied evaluation model provides no solution for assessment of such MCQ unit types. Hence, we analyse further only 3424 content-based MCQ units based on 1038 texts.

\subsubsection{Variety of errors} \label{sec:varerr}
As a corpus of reading comprehension MCQs for EFL learners, RACE contains issues that are generally avoided by reading comprehension test developers ``because of possible flaws and/or lack of text dependency of a question'' \citep{haladyna2004developing}. These problems include language and formatting errors, incomplete MCQ elements, MCQ units considered inadequate specifically for a reading comprehension test. 

More specifically, all MCQs from the RACE corpus are evaluated and classified in terms of problems related to the text, question, alternatives (or a combination of these). Each problematic MCQ element, i.e. text, question, alternative(s), is analysed according to the four problem categories, depending on the degree of negative impact an issue (potentially) has on the reading comprehension process:
\begin{itemize}
    \item \textbf{unacceptable} \textit{(severe problems)} – either prevents a reader from understanding the affected piece of information or changes it in a way that content becomes unclear or ambiguous;
    \item \textbf{partially acceptable} \textit{(moderate problems)} – the affected piece of information contains one or more spelling errors while the correct word/meaning can certainly be reconstructed from another part of the same MCQ element, and/or the affected MCQ element contains information which is not (anymore) related to the given content or is unnecessary for a reader in the context of a reading comprehension test;
    \item \textbf{mainly acceptable} \textit{(mild problems)} – certain punctuation rules are not followed in the affected piece of information;
    \item \textbf{acceptable} \textit{(no problems)} – all MCQ elements are free from the problems viewed here as mild, moderate, or severe.
\end{itemize}
The classification above applies to all MCQ elements and their possible combinations.
A complete typology of the described problems is provided in Table \ref{tab:errortypes}. For definitions, see Appendix \ref{app:B}; for examples, see \cite[Section 3]{zyrianovam}.

Additionally, the texts in RACE appear to have used different \textbf{\textit{varieties of English}} (British English and American English), where lexical (e.g., \textit{biscuits - cookies}, etc.), grammatical (e.g., \textit{a family love cooking - a family loves cooking}, etc.), and spelling (e.g., \textit{colour - color, prise - prize}, etc.) differences are not avoided. The issue was not covered by \cite{lai-etal-2017-race}. In this article, British and American varieties of the English language are controlled in accordance with corresponding dictionaries and rules of British English (Collins English Dictionary\footnote{Access at: https://www.collinsdictionary.com/dictionary/english}; \cite{Biber_99} and American English (Merriam-Webster Dictionary\footnote{Access at: https://www.merriam-webster.com/}; Straus, Kaufman, Ster (2014)). We do not provide any statistics on this matter, but rather consider erroneous language use for British/American English as a spelling error, as described in Appendix \ref{app:B} (Glossary of errors).

\begin{sidewaystable}[]
\caption{Typology of errors}
\label{tab:errortypes}
\begin{tabular}{@{}ll|ll|ll@{}}
\toprule
\multicolumn{2}{c}{\textbf{Severe}} & \multicolumn{2}{c}{\textbf{Moderate}}                 & \multicolumn{2}{c}{\textbf{Mild}}         \\ \midrule
                        &           & \multicolumn{1}{c}{\textbf{}}             & \textbf{} & \multicolumn{1}{c}{\textbf{}} & \textbf{} \\
Incomplete   text       & T         & Spelling errors (hyphens,   contractions) & T, Q, A   & Extra spaces (punctuation)    & T, Q, A   \\
Misleading   gaps       & T, Q      & Additional notes                          & T, Q, A   & Missing spaces (punctuation)  & T, Q, A   \\
Extra   gaps                                        & T, Q    &                         &   & Punctuation errors       & T, Q, A \\
Misleading   spaces                                 & T, Q, A &                         &   & Formatting inconsistency & T, Q, A \\
Extra   spaces (within a word or a digit)           & T, Q, A &                         &   &                          &         \\
Missing   spaces (between words and/or digits)      & T, Q, A &                         &   &                          &         \\
Misleading   spelling errors                        & T, Q, A &                         &   &                          &         \\
Spelling   errors (except hyphens and contractions) & T, Q, A &                         &   &                          &         \\
Grammatical   errors                                & T, Q, A &                         &   &                          &         \\
Syntax   errors                                     & T, Q, A &                         &   &                          &         \\
OCR   errors                                        & T, Q, A &                         &   &                          &         \\
Time-dependent                                      & T, Q, A &                         &   &                          &         \\
                                                    &         &                         &   &                          &         \\
Incomplete   question                               & Q       &                         &   &                          &         \\
Answerable   without reading                        & Q       &                         &   &                          &         \\
Subjective   formulation                            & Q, A    &                         &   &                          &         \\
Ambiguously   formulated                            & Q       &                         &   &                          &         \\
                                                    &         &                         &   &                          &         \\
Incomplete   alternative(s)                         & A       & Inconsistency between A & A &                          &         \\
Overlapping   alternatives                          & A       &                         &   &                          &         \\
                                                    &         &                         &   &                          &         \\
                                                    &         &                         &   &                          &         \\
Inconsistency   Q and A                             &         &                         &   &                          &         \\
Inconsistency   Q and T                             &         &                         &   &                          &         \\
Inconsistency   T and A                             &         &                         &   &                          &        
\end{tabular}
\end{sidewaystable} 

\begin{figure}[t!]
    \centering
    \includegraphics[width=0.61\textwidth]{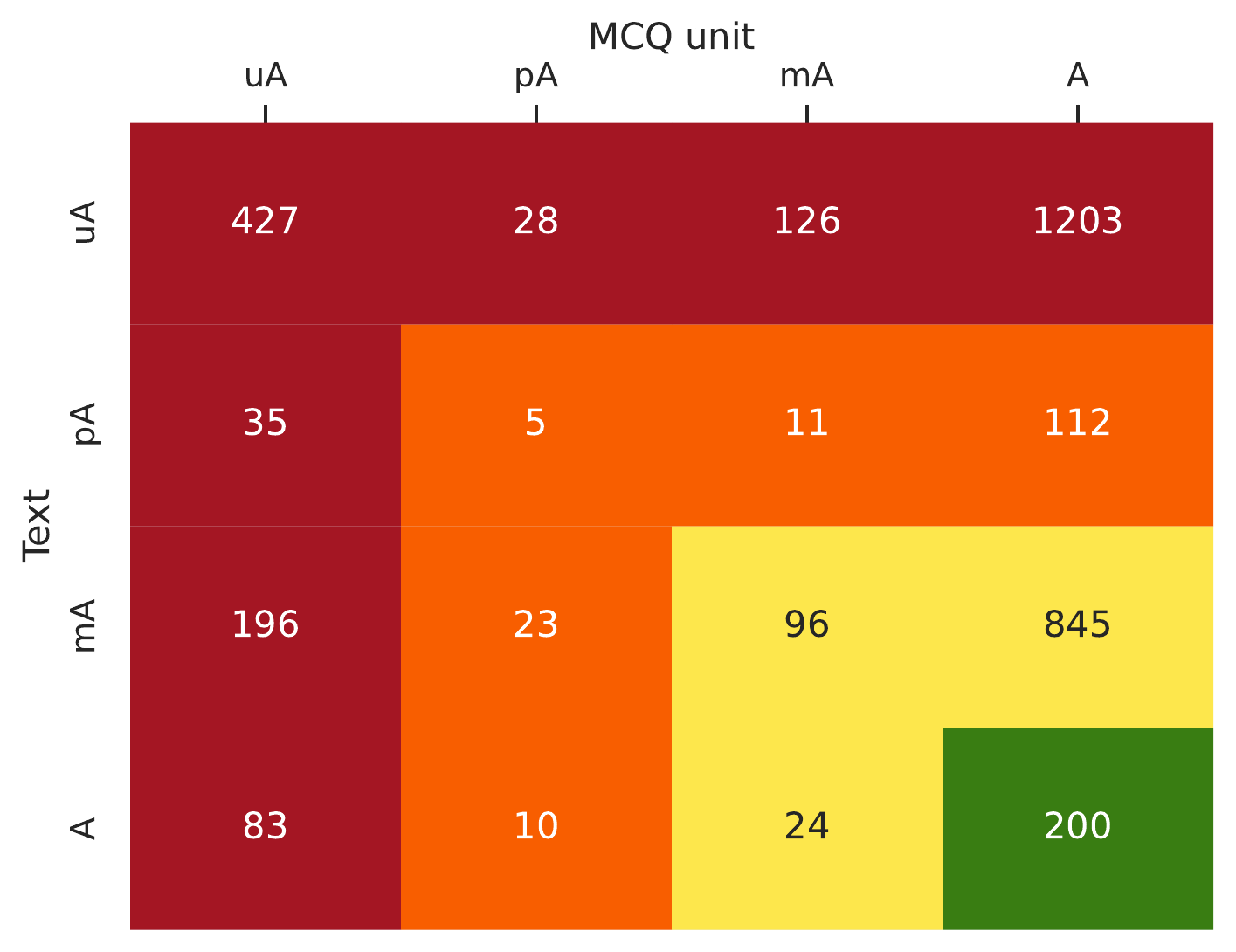}
    \caption{Distribution of MCQs per problem category. uA stands for ``unacceptable'', pA means ``partially acceptable'', mA denotes ``mainly acceptable'', A means ``acceptable''}\label{fig:heatmap}
\end{figure}

The distribution of the 3424 MCQs per the aforementioned problem categories for text and MCQ unit is shown in Figure  \ref{fig:heatmap}. We note that only 200 of 3424 MCQs contain no problematic issues, neither in the passages nor in the questions or alternatives, and are treated as acceptable. We also note that 2160 (63\%) of MCQs (the right-most column in Figure \ref{fig:heatmap} except for the green cell) cannot be classified as acceptable \emph{only} due to the problems related to the texts. Furthermore, more than half of these (1203) include texts with severe problems. Together with that, there are also 83 MCQs with acceptable texts but severe problems in the question and/or alternatives. 

\textbf{Note:} one MCQ element might contain problems of different types, in such cases only the one making the most negative impact is included in Figure \ref{fig:heatmap}. If one and the same text contains certain problem(s) and has more than one related MCQ unit, each of these MCQs is affected.

Recall that out of 1045 texts of the RACE corpus, only 1038 are included at this stage of the analysis (for the reasons to exclude the others, see sections \ref{sec:vartext}, \ref{sec:varmcq}). Among the 1038 analysed passages, only 92 are acceptable (those are texts of MCQs in the bottom row in Figure \ref{fig:heatmap}) and 946 contain text-related problems of various types.

\begin{figure}[!t]
	\centering
	\includegraphics[width=0.97\linewidth]{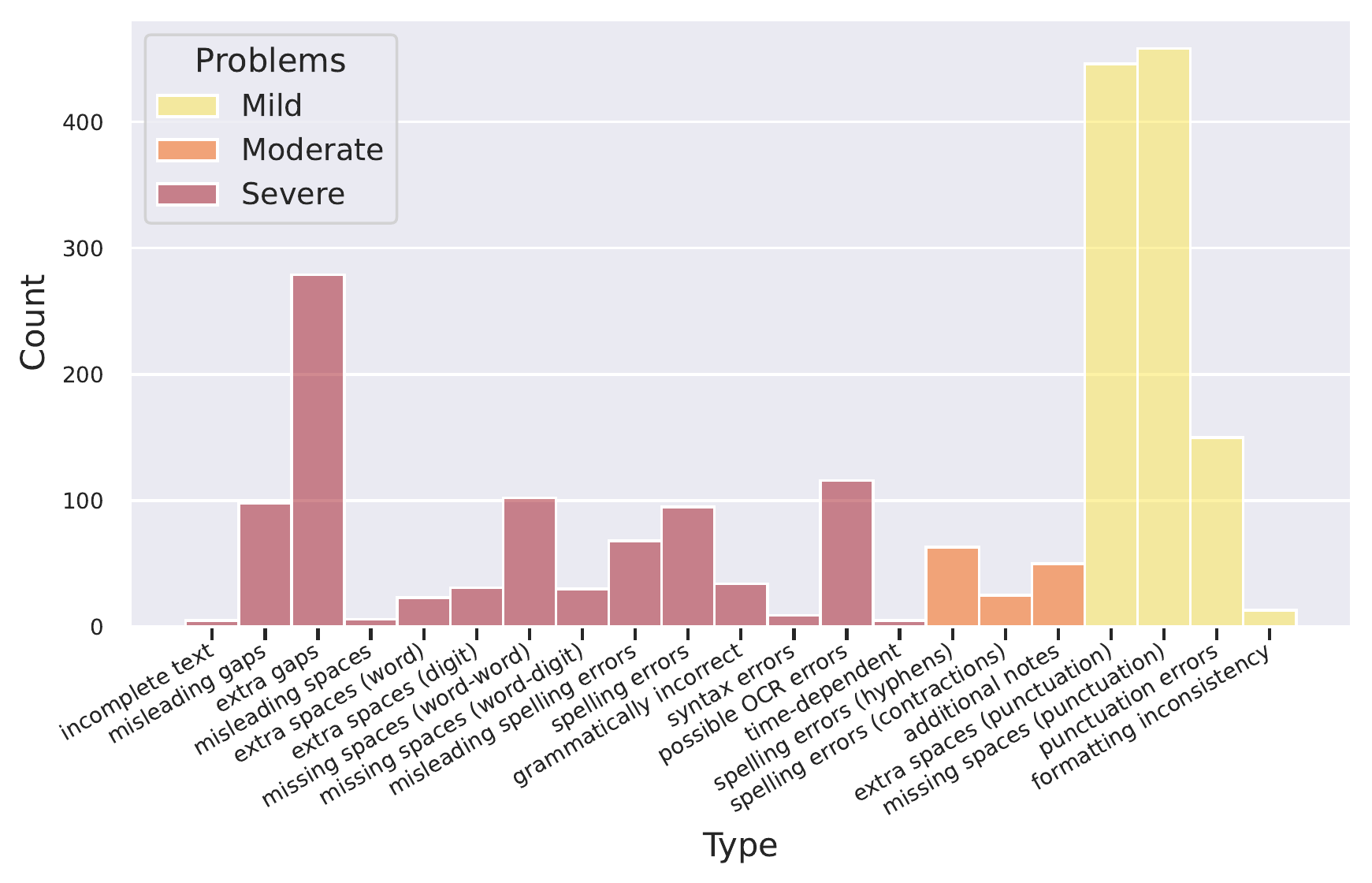}
	\caption{Distribution of problems in texts}
	\label{fig:textproblems}
\end{figure}

\begin{figure}[!t]
	\centering
	\includegraphics[width=0.97\linewidth]{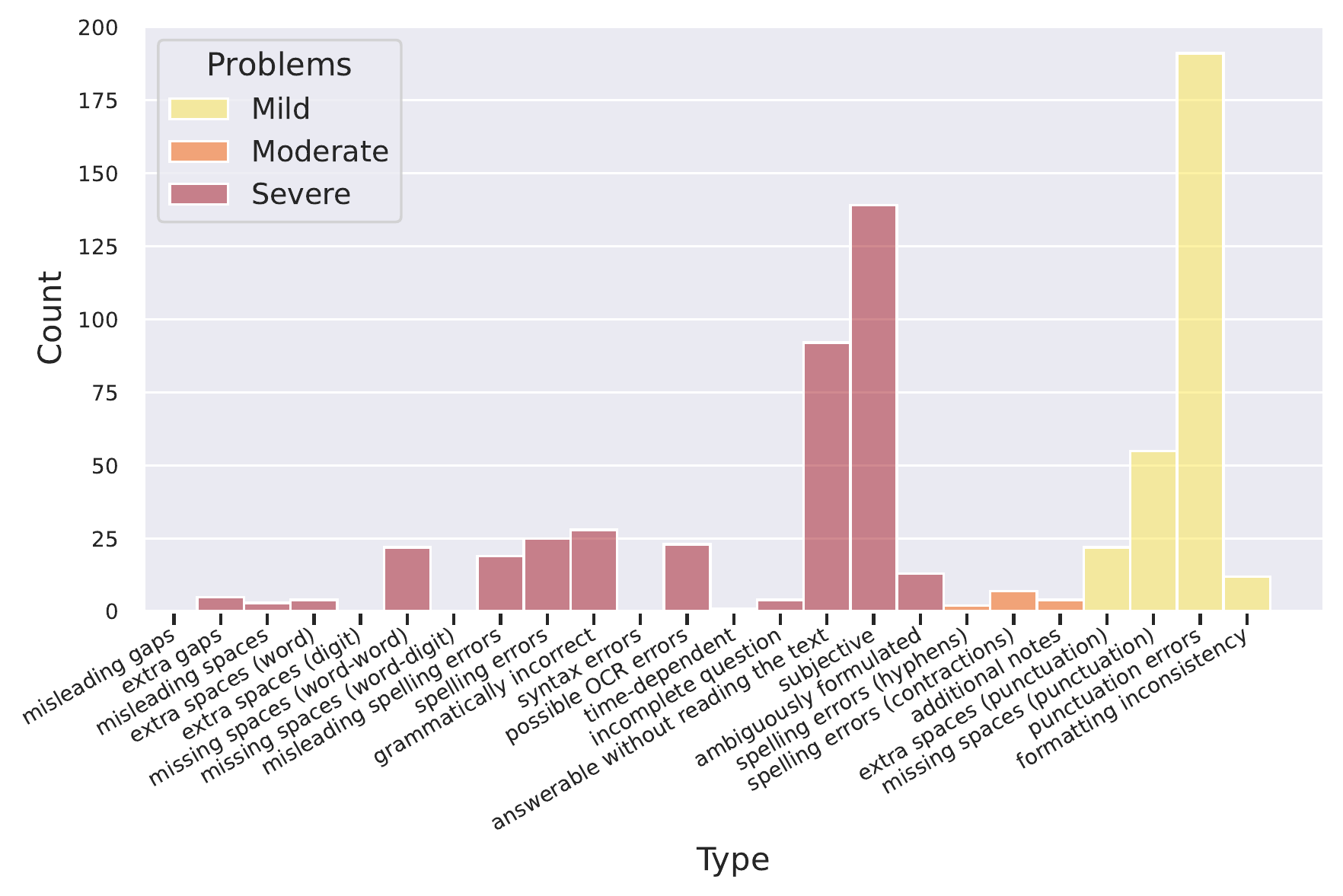}
	\caption{Distribution of problems in MCQ stems}
	\label{fig:stemproblems}
\end{figure}

\begin{figure}[!t]
	\centering
	\includegraphics[width=0.97\linewidth]{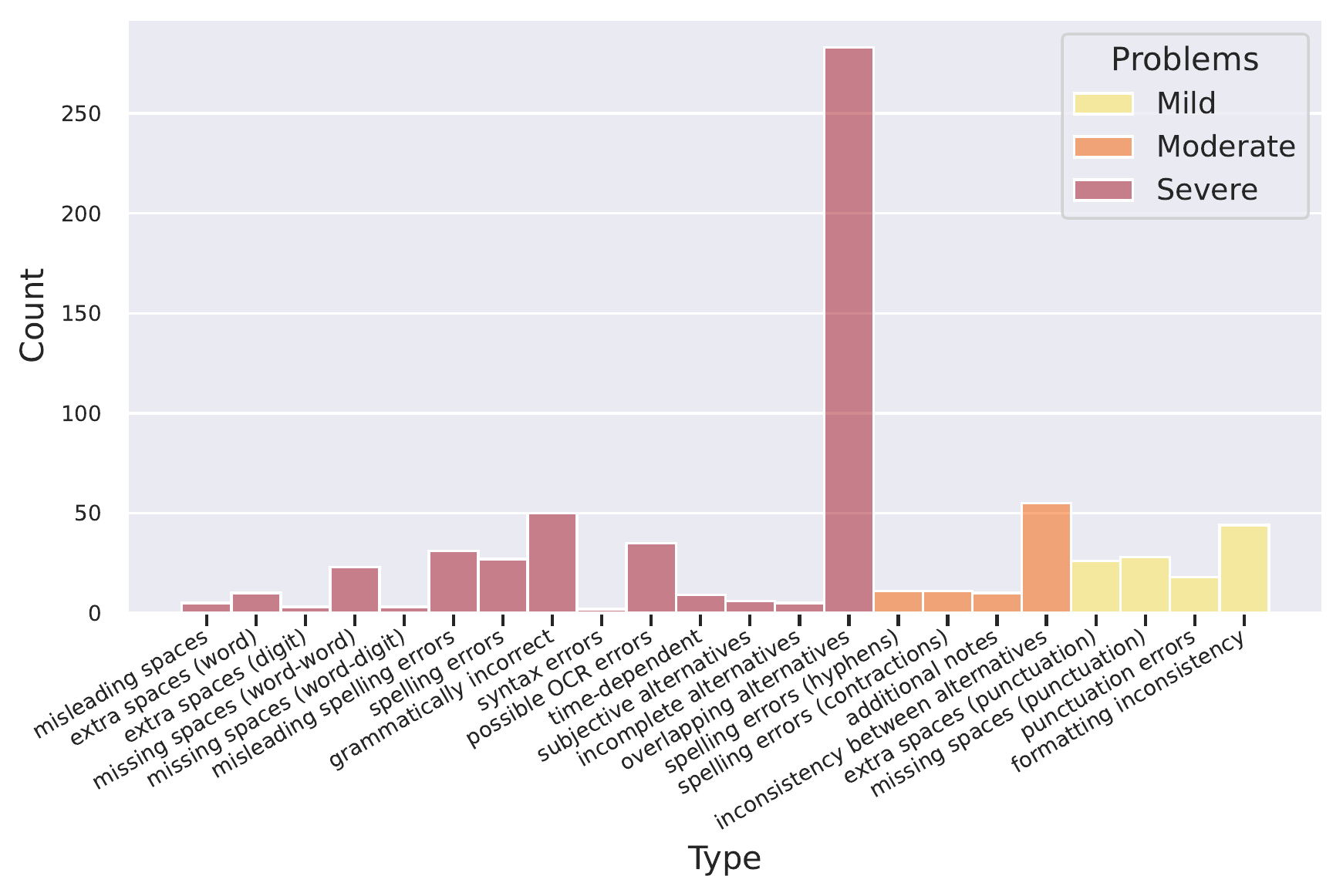}
	\caption{Distribution of problems in MCQ alternatives}
	\label{fig:choicesproblems}
\end{figure}

\begin{figure}[!t]
	\centering
	\includegraphics[width=0.7\linewidth]{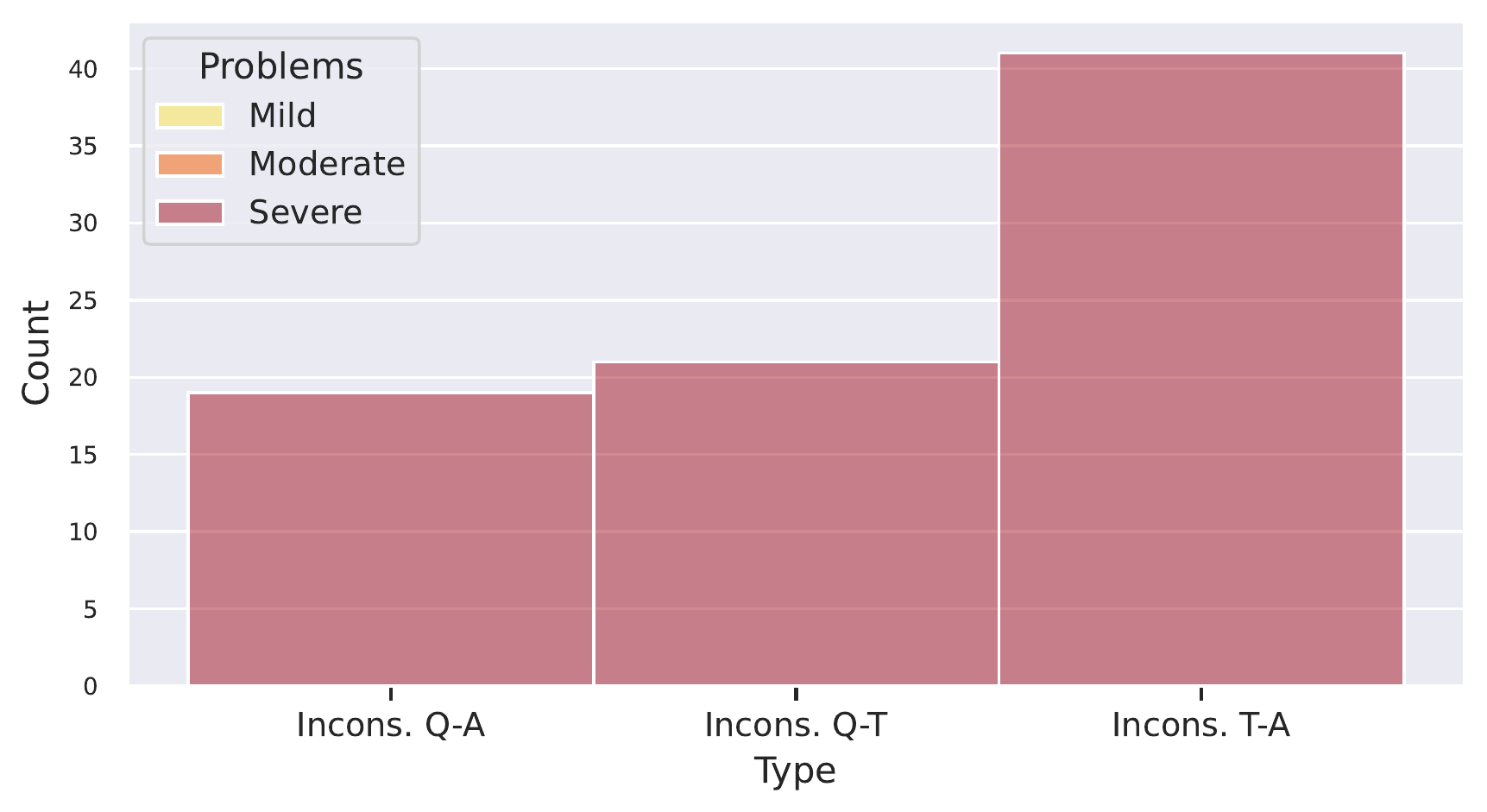}
	\caption{Distribution of problems in the interaction between the text and MCQ elements}
	\label{fig:interactionproblems}
\end{figure}

Distributions of the MCQ-related problems identified in texts, questions, alternatives, and their possible combinations are provided in Figure \ref{fig:textproblems}, Figure \ref{fig:stemproblems}, Figure \ref{fig:choicesproblems}, and Figure \ref{fig:interactionproblems}, respectively. For Figure \ref{fig:choicesproblems} the tiers (i.e. mild, moderate, severe) relate to the quality of alternatives as a whole rather than to each of them separately.

As seen from the Figures \ref{fig:textproblems}, \ref{fig:stemproblems}, \underline{the most frequent type(s) of problems} are mild problems both in texts and questions (458 MCQs with \emph{missing} and 446 MCQs with \emph{extra spaces} in the texts; 191 MCQs with \emph{punctuation errors} in the questions). However, when it comes to alternatives (Figure \ref{fig:choicesproblems}), the highest number of the problems (which is substantially larger than all the others) is related to \emph{overlapping alternatives} (283 MCQs), while the second highest is \emph{inconsistency between alternatives} (55 MCQs), considered severe and moderate problems, respectively (see \ref{sec:varerr}). Among the problems related to the combinations of the MCQ elements (Figure \ref{fig:interactionproblems}), \emph{inconsistency between the text and alternatives} turns out to be the most frequent one (41 MCQs).

Interestingly, the same kinds of problems were reported to be frequent in previous works on distractor generation models, e.g., by \cite{kalpakchi-boye-2021-bert}. Given that the alternatives in RACE were generated by humans, as pointed out by the RACE developers \citep{lai-etal-2017-race}, the degree to which one can rely on human annotators when judging the quality of distractors needs to be investigated further.

\subsection{MCQ difficulty} \label{sec:mcqdiff}
In this section we analyse only the MCQs where both passages and MCQ units are acceptable (contain no problems), mainly or partially acceptable (contain mild or moderate problems) -- 1326 MCQs altogether based on 466 texts. MCQs with severe problems in any of the MCQ elements are not taken into consideration for difficulty evaluation since these problems (see Section \ref{sec:varerr}, Table \ref{tab:errortypes}) might influence the evaluation, (potentially) making the final score incorrect or irrelevant.

Additionally, we exclude all MCQs with partly continuous texts (145 MCQs), since we have no evidence of the kind of impact elements of a non-continuous text (e.g., \textit{(sub)titles, sections fixed for all parts of a multiple-text unit}, etc.) do make on the MCQs difficulty. This issue is not covered by the evaluation model developed by \cite{kirsch1995interpreting}, \cite{kirsch1999}, \cite{kirsch2001international}, and as far as the model is in the core of this research, we cannot reassure that the final score for MCQs related to the partly continuous texts is as precise as it is for the purely continuous texts. The remaining \emph{1181 MCQs (on 412 texts)} are subject to the difficulty analysis.

\begin{figure}[!b]
	\centering
	\includegraphics[width=\linewidth]{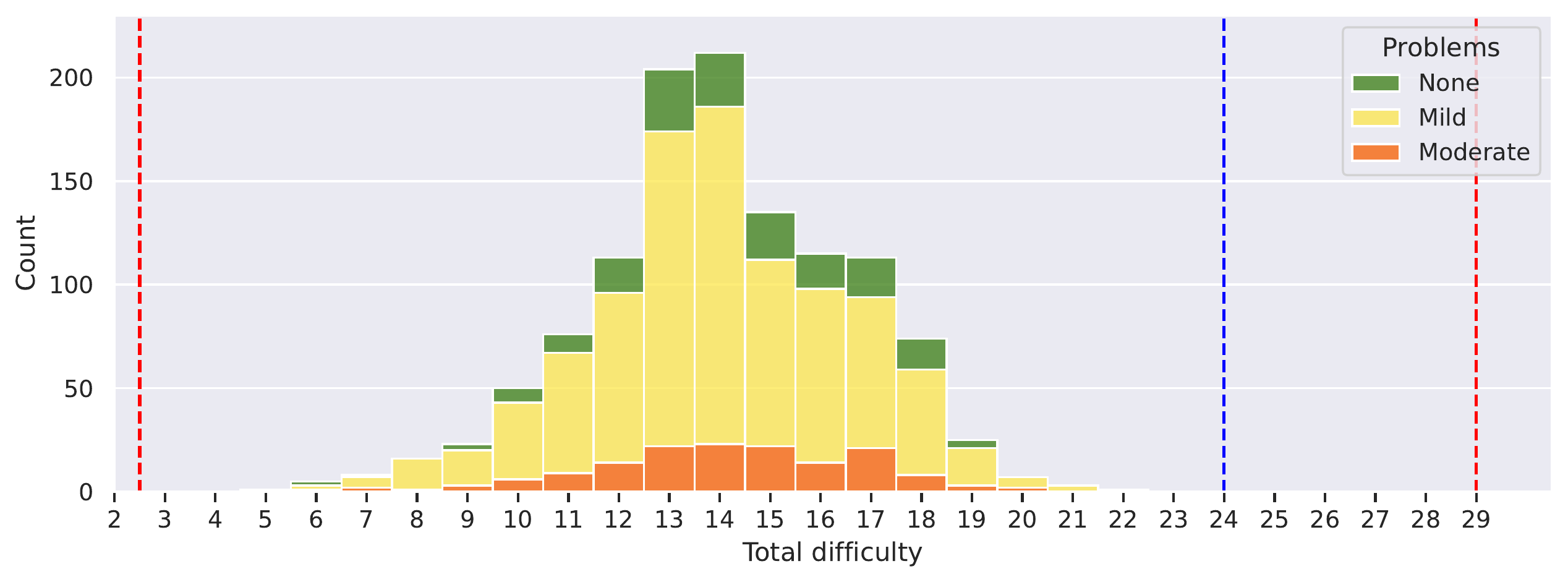}
	\caption{Distribution of the total difficulty for the evaluated MCQs. Each bin represents a half-open interval, e.g., $[8, 9)$ for the bin centered around 8. The red dashed lines represent theoretical min. and max. values for the total difficulty. The blue dashed line represents a theoretical maximum without TOC.}
	\label{fig:ok_diff_dist}
\end{figure}

Distribution of the \textbf{\textit{total difficulty}} for the evaluated MCQs is visualised in Figure \ref{fig:ok_diff_dist}, where the more difficult a MCQ is, the more points it scores, according to the applied evaluation scheme (see Section \ref{sec:varcond}, Appendix \ref{app:A}). We observe a slightly skewed normal distribution of difficulty scores (mode of 14, median of 13, and mean of 13.09). We note that the observed mean and median differ from the theoretical mean and median (both 15.75). However, bearing in mind that theoretical maximum includes scoring maximum on TOC, which is not supposed to be common in RC tasks, we look at the relaxed theoretical maximum of 24 without TOC (blue dashed line). Then the theoretical mean and median are brought down to 13.25, which makes the difference with the observed ones much less substantial. The observed differences are due to having no MCQs with difficulty less than 5 or more than 22 in the RACE corpus.

The central variables to the total difficulty are Type of match (TOM), Type of information (TOI), and Plausibility of distractors (POD). These 3 variables together (out of 9 in total) account for 51.7\% of possible difficulty points (or 62.5\% excluding the less influential TOC variable). Below we analyse each of these central variables separately.

\begin{figure}[!t]
	\centering
	\includegraphics[width=\linewidth]{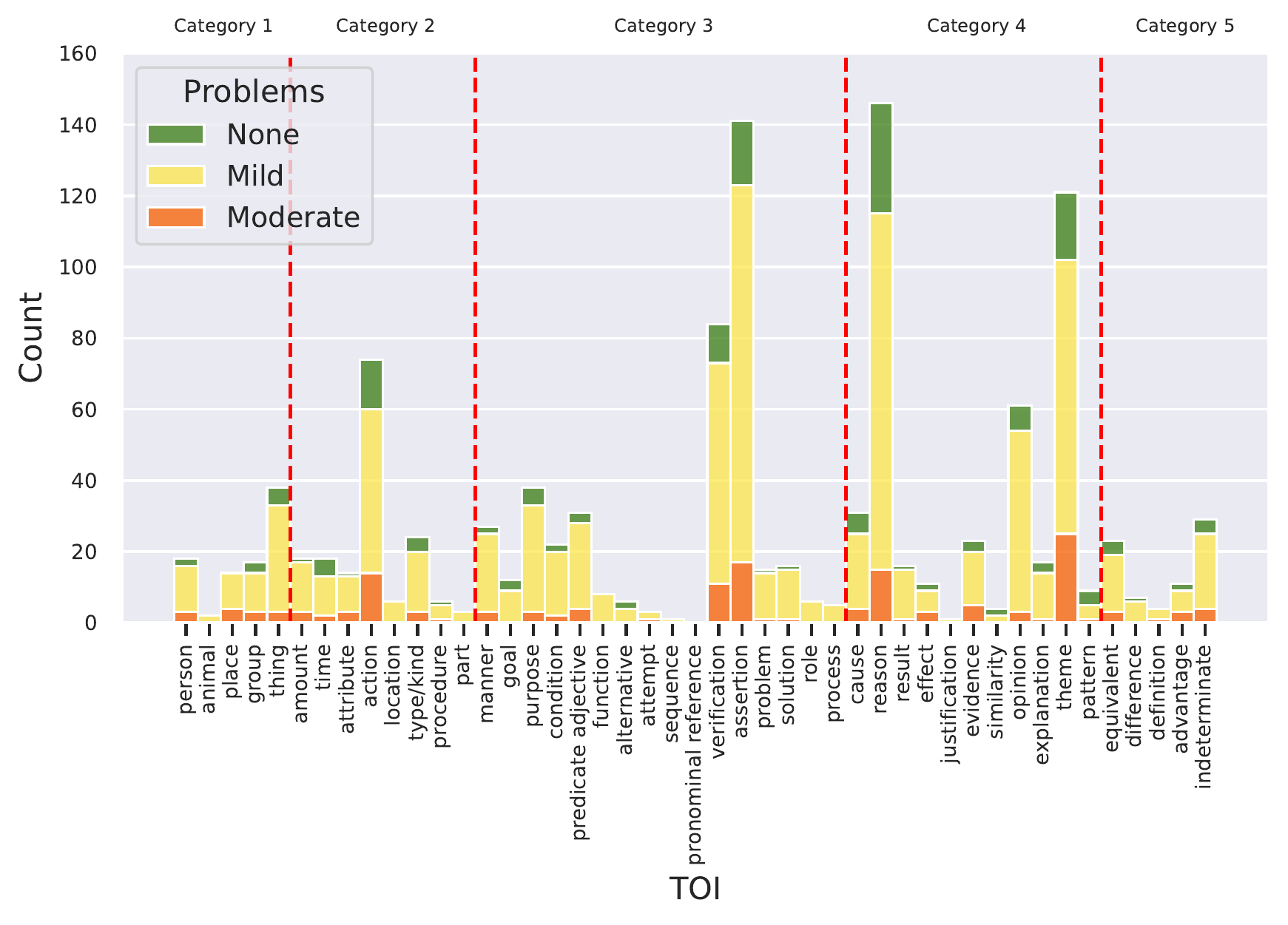}
	\caption{Distribution of the TOI concepts for the evaluated MCQs. The concepts are ordered from the lowest to the highest scoring (minimum of 1 and maximum of 5). The red dashed lines signify the increment of the TOI score by 1.}
	\label{fig:ok_toi_dist}
\end{figure}

As seen from the Figure  \ref{fig:ok_toi_dist}, the distribution of the MCQs from the perspective of \textbf{\textit{TOI}} is not balanced within the corpus: while there are more than 120 MCQs for \textit{reason, assertion, and theme} each, there are concepts like \textit{animal, location, procedure, part, function, alternative, attempt, sequence, role, process, justification, similarity, difference, or pronominal reference}, which appear underrepresented in the corpus, with less than 10 corresponding MCQs per each concept.

In addition, for certain concepts (e.g., \textit{place, attribute, function, problem, solution, result}, among others) there are no acceptable MCQs, though the majority of them are mainly acceptable (contain only mild problems). We also note that five largest TOI concepts with MCQs having moderate problems (orange in Figure \ref{fig:ok_toi_dist}) are \textit{theme, assertion, reason, action, and verification}. The distributions within categories do not appear to have any structure or follow any theoretical distribution. Additionally, all categories have substantial skews towards one or two TOI concepts.

\begin{figure}[!t]
	\centering
	\includegraphics[width=0.94\linewidth]{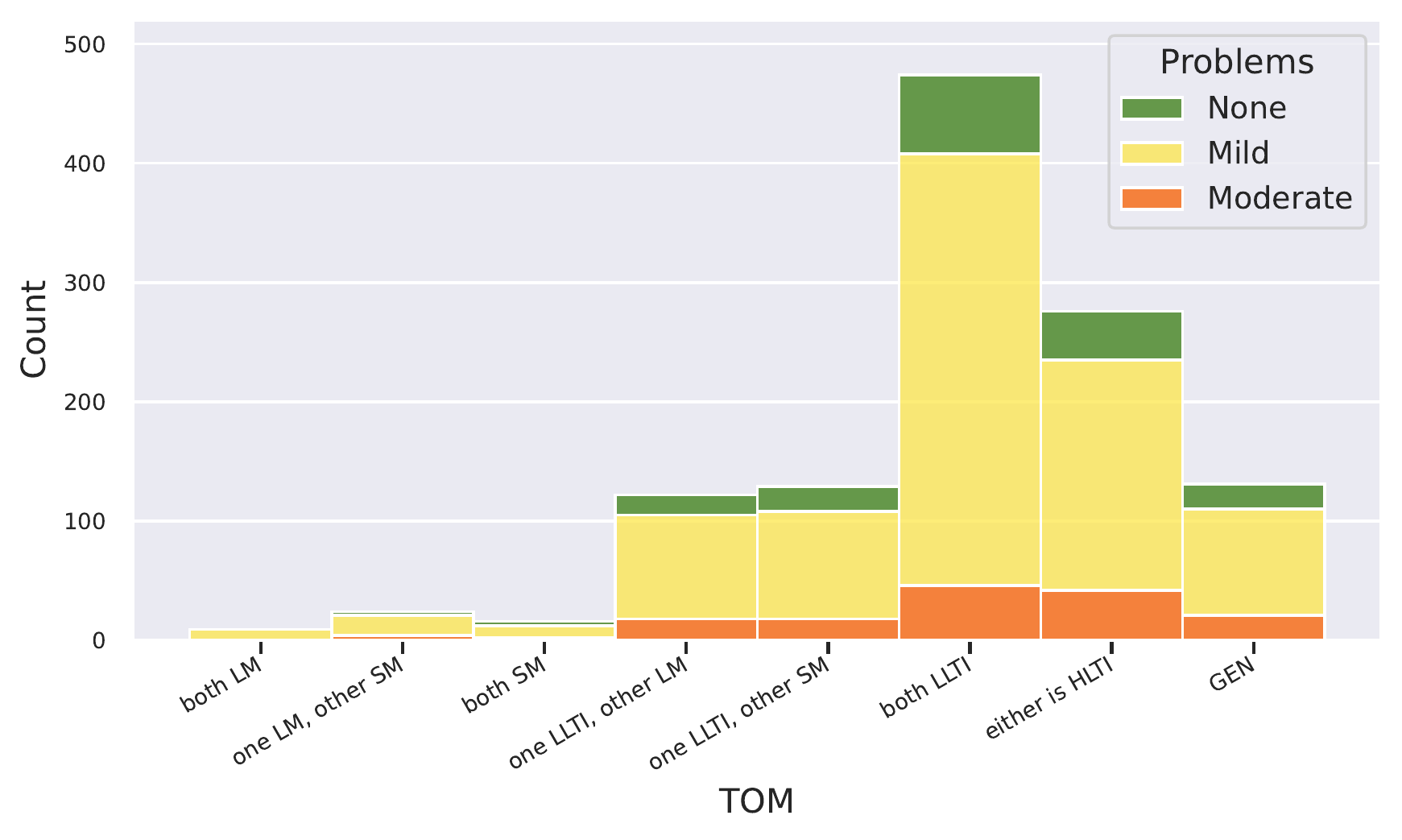}
	\caption{Distribution of the TOM categories for the evaluated MCQs. The categories are ordered by their scores with the lowest scoring one being the leftmost.}
	\label{fig:ok_tom_dist}
\end{figure}

The distribution of MCQs with respect to \textbf{\textit{TOM}} is presented in Figure \ref{fig:ok_tom_dist}. We observe that the number of MCQs where both T-Q and T-A relations are represented by low-level text-based inferences is substantially higher (474 MCQs) than that of any other category. Together with that, there is a noticeable drop in MCQs based on literal or synonymous match (or a combination of these) -- three respective categories taken together result in less than 100 MCQs in total, among the 1181 evaluated MCQs.

\begin{figure}[!ht]
	\centering
	\includegraphics[width=0.94\linewidth]{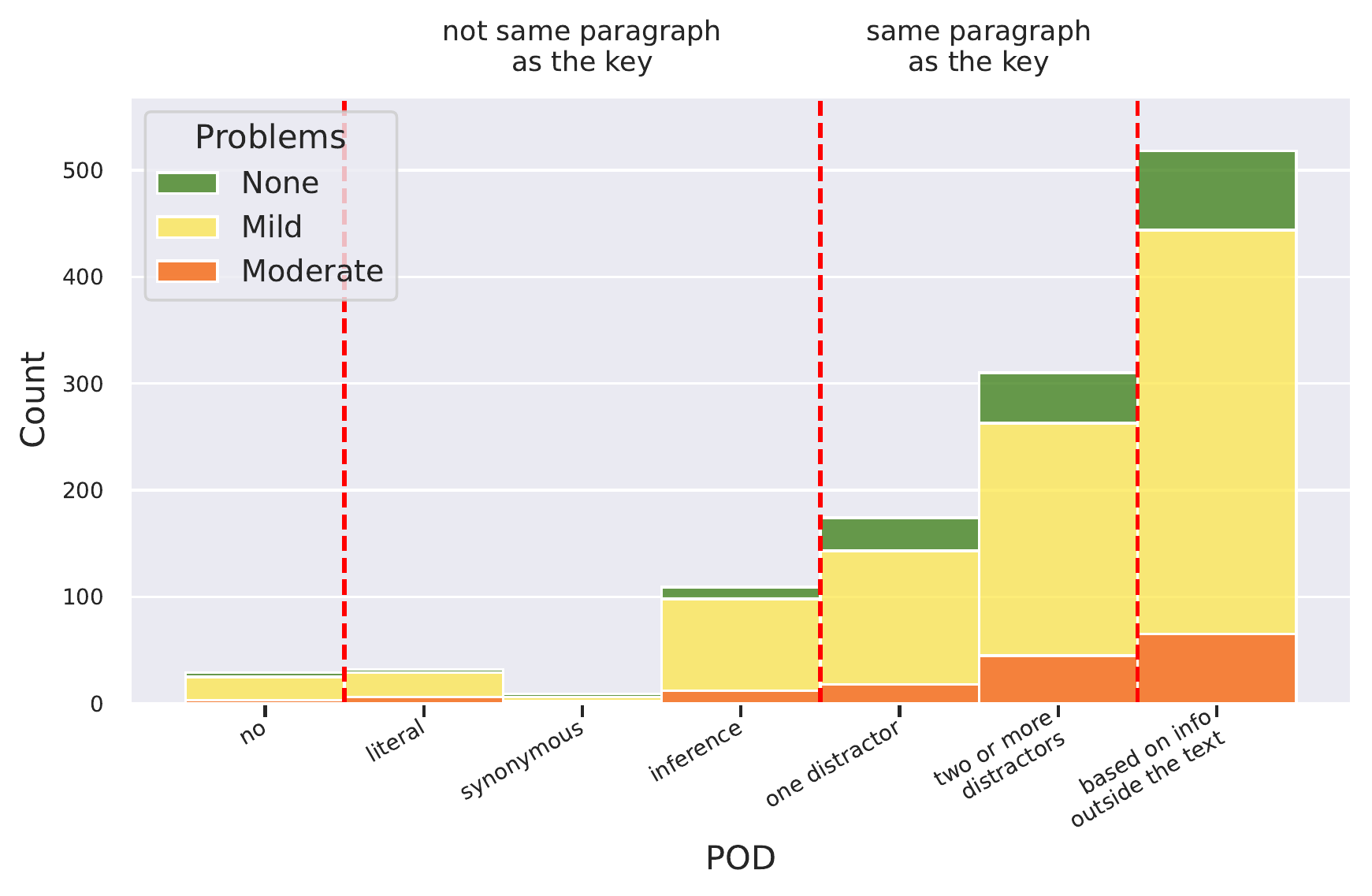}
	\caption{Distribution of the POD categories within the evaluated MCQs. The categories are ordered by their scores with the lowest scoring one being the leftmost. The categories scoring the same are ordered following \cite{kirsch1995interpreting}}
	\label{fig:ok_pod_dist}
\end{figure}

The distribution of \textbf{\textit{POD}} scores for the evaluated MCQs, shown in Figure \ref{fig:ok_pod_dist}, is close to exponential, meaning the higher the score, the more MCQs get it. We observe that MCQs with at least one distractor being based on information outside the text are the most frequent (518 MCQs). The runner-up category includes MCQs where two or more distractors (or corresponding bases) grounded on information from the text appear in the same paragraph(s) with the key (310 MCQs). Both categories score the theoretically maximal 5 points for POD.

However, we note that the distribution of the POD categories is not equal either. Here, distractors which are synonymous to the corresponding bases from the text but appear in a different paragraph(s) from the key, are the least represented. At the same time, the number of MCQs with no distracting information is close to the number of cases when the distractors match literally to the text, but are not in the same paragraph as the key.

\subsection{Bases for alternatives} \label{sec:addan}
When annotating the MCQs from the RACE corpus, we additionally mark bases for alternatives. Here, \textit{basis} is a a piece of information from the text which allows to make a corresponding match or inference between such piece of text and an alternative. However, the basis cannot always be marked since some alternatives are based on information not directly stated in the text (i.e. \textit{keys} which imply generating an appropriate interpretative framework -- see Section \ref{sec:tom};  \textit{distractors} which are based on information outside the text -- see Section \ref{sec:pod}). When possible, though, bases for these alternatives are also marked as a way to relate a key/distractor to the text. 

\begin{figure}[!b]
	\centering
	\includegraphics[width=\linewidth]{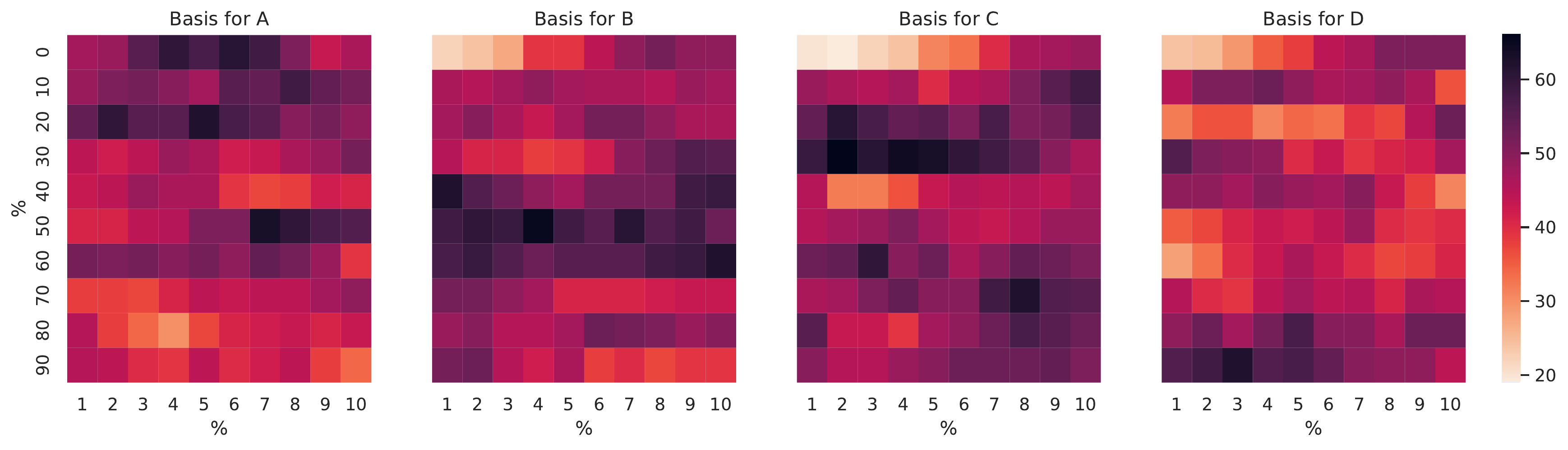}
	\caption{Distribution of the bases' positions in the text. To exemplify how to read the chart: to find a bucket that corresponds to 34\% of characters in the text, find the row with the label ``30'' first and then the column with the label ``4''. The cell at their intersection will indicate the 34\%-bucket.}
	\label{fig:ok_bases_dist}
\end{figure}

In Figure \ref{fig:ok_bases_dist} we visualise the distribution of bases' positions in the text for each alternative. For this analysis, the same 1181 MCQs as in section \ref{sec:mcqdiff} are taken into consideration. Each text is divided into character buckets such that one bucket constitutes 1\% of characters in the text (represented by one square in each subfigure of Figure \ref{fig:ok_bases_dist}). Hence, any text is divided into 100 buckets (all squares together). For each basis, we have marked all buckets between the starting and finishing character of the basis. To get a heatmap in Figure \ref{fig:ok_bases_dist} for one alternative, say A (the leftmost plot in Figure \ref{fig:ok_bases_dist}), we superposition the bases' buckets for alternative A in each text of for all MCQs under consideration.

As seen from the visualisation, the distribution of the bases' positions is not equally balanced for the analysed MCQs. Bases for an alternative A are more frequently located at the beginning of the texts (up to 30\% character-wise). Similarly, bases for an alternative D are found at the end of a text more often than in any other part. This kind of bias is not necessarily desirable while evaluating the models for both MCQ answering and, especially, generation. To exemplify the latter case, consider a model prone to generating alternative A based on information at the beginning of the text, and alternative D based on the information at the end, no matter if the information is actually relevant to the stem. If evaluated on such data, especially using only automatic evaluation metrics, the model might perform quite well because alternative A happened to be most often at the beginning of the texts (and D at the end). Although, clearly the model that does not take the stem into account, is not a good MCQ generation model. Hence, we strongly recommend against using automatic evaluation metrics on existing MCQs when evaluating generation models (particularly on the test set for RACE-H).

\section{Discussion}\label{sec:disc}
The evaluation model applied to the RACE corpus has its limitations. One limitation is that it does not cover the following aspects which might influence MCQ difficulty, especially for EFL learners:
\begin{itemize}
    \item \textbf{when evaluating TOM}
       
    \begin{itemize}
        \item \textit{Q-A relations} do not impact the final difficulty score, although it might be necessary, for instance, when making a pronominal reference between the stem and one or more alternatives, consequently;
        \item the process of selecting an exact piece of corresponding information from the text (\textit{the basis}) for each alternative is highly context-dependent, potentially influencing both the number of required paragraphs and the relative location of bases for the alternatives, having impact on the TOM and/or NPar scores;
        %- information from the Q might or might not be included to the marked basis;
        \item MCQs where \textit{either T-Q or T-A relations} require a HLTI, while another one is based on a LM, or SM, or LLTI, are not differentiated and score the same, in contrast to other cases where LM, SM, and LLTI are part of TOM;
        \item \textit{for SM}, the number of substitutions from literal to synonymous correspondence does not influence the TOM score, hence there is no difference in score between substituting one word/phrase for synonyms, or four.
    \end{itemize}
        
    \item \textbf{for NPar}, all MCQs which require two, three, or more paragraphs score the same;
    \item \textbf{the NPar variable} does not obligatorily reflect the exact amount of information that a reader should process (e.g., only one sentence might be necessary from an additional paragraph to make a pronominal reference, though the whole paragraph is marked as required, regardless of its length, format, or other characteristics);
    \item since \textbf{NPar and IC} are connected, IC of contrasting does not score additional points when MCQs require two or more paragraphs to make a correct inference.
\end{itemize}

Apart from the variable-related cases discussed above, the model does not consider \textit{readability dimensions} like number of syllables/words/punctuation marks etc. per \textit{n} words/clauses/sentences/paragraphs. \cite{kirsch1999} reported the dimension (based then on the average number of syllables per 100 words and the average number of sentences per 100 words) as not significant (for native English readers) when combined with the process variables of TOI, TOM, POD. However, the importance of including readability when evaluating MCQ difficulty for EFL learners is yet to be investigated.

It should also be noticed that the applied evaluation model appears highly sensitive to the exact formulation of a stem, which means that questions might be similar in nature but score differently depending on the exact formulation alone. This includes (but is not limited to) the \textit{structure of a stem} (interrogative sentence or fill-in-the-gap task) which might influence NPhr; cases when \textit{nouns or pronouns are (not) used in the stem} and cases when \textit{additional information from the text is (not) included to the stem} which might influence NPar. Similarly to stems being context-dependent, MCQ units are viewed as context-dependent, since it is their particular relation to a given text that is used to define relevant categories for other variables.

\section{Related work}\label{sec:relw}
\subsection {Quality evaluation of RACE}
RACE has been widely used as a dataset to evaluate machine reading comprehension models, though the dataset was rarely analysed in terms of its structure and language quality.  
Nevertheless, when presenting RACE-C as a multi-choice reading comprehension dataset, \cite{liang2019new} pointed out that  (1) not all questions that contain keywords ``underline'' or ``underlined'' had been removed from RACE, although removing such question was one of the filtering steps described in \cite{lai-etal-2017-race}, and (2) the samples in RACE are duplicated to a certain extent. In addition, \cite{berzak-etal-2020-starc} demonstrated that 18\% of MCQs were unanimously judged by humans as not having a unique correct answer, while 47\% of RACE questions can be guessed by machines without accessing the passage. A similar concern was expressed by \cite{zou2021palrace} regarding RACE for high-school (RACE-H). With a reference to previous studies by \cite{berzak-etal-2020-starc} and \cite{si2019does} respectively, they claimed that some questions in RACE have wrong answers and some can be guessed without reading passages. After examination on these issues, only 800 MCQs passed the verification and some typos were then also corrected within the mentioned research.

Some scholars pay additional attention to the typology of the passages and multiple-choice questions included to RACE. Thus, \cite{xu-etal-2022-fantastic} note that in the dataset, there are both narrative and informational paragraphs with no specification of the genre added to the texts. According to \cite{liebfreund2021cognitive} (as cited by \cite{xu-etal-2022-fantastic}), these genres require different comprehension skills, and mixing them ``obscures the ability to offer a precise assessment'' \citep{xu-etal-2022-fantastic}. \cite{sun-etal-2019-improving} remark that 82.0\% of the MCQs in RACE require one or multiple types of world knowledge, where ``world knowledge'' can refer to negation resolution, common sense, paraphrase, and mathematical/logic knowledge. \cite{wang2019evidence} also state that ``even humans have troubles in locating evidence sentences'' (i.e. sentences which represent the correct answer in the passage), when questions require understanding of the entire document and/or external knowledge. According to them, slots for 10.8\% of total questions were left blank by at least one annotator due to these challenges.

\subsection{Operationalisable MCQ difficulty evaluation}
Difficulty of multiple-choice questions for reading comprehension is commonly measured after readers complete the test by proportion of the readers who give correct answers to those who fail the task \citep{kirsch1995interpreting, cook1988comparative, katz1990answering}. These approaches usually have no specification if to be applied to test native speakers' skills or those of a foreign language learners. However, it is stated by \cite{oecd2019pisa} that non-native speakers of the target language might be relatively slow readers and yet possess compensatory or strategic processes that permit them to be higher-level readers when given sufficient time to complete complex tasks. Hence, a specific framework to evaluate reading skills of learners of English as a foreign language is proposed by \cite{oecd2021pisa}. The assessment framework is based on descriptors and proficiency levels designed for the CEFR (Common European Framework of Reference for Languages: Learning, Teaching, Assessment). The proficiency levels allow to describe the functions a learner can or cannot do at a certain stage of the language acquisition process, while the framework stays hardly operationalisable in terms of a scoring system to apply.
    
There are also scholars who suggest evaluating parts of a multiple-choice question (i.e. given text, question, distractors, correct answer) separately, by involving different dimensions. These can include, among others, readability of the text, question, and the correct answer \citep{Freedle1993}, vocabulary and syntactic complexity of texts \citep{Park2012}, number and order of distractors \citep{Rodriguez2005}, location of the most attractive distractor \citep{Shin2020}.

\section{Conclusion}\label{sec:conc} 
Despite RACE being constructed by humans and widely used for evaluating MCQ answering and generation models, our analysis of the test set of RACE-H demonstrates that it does not entirely fulfill the basic requirements for MCQs. Furthermore, the corpus contains certain biases in positions of the bases for the alternatives in corresponding texts. Consequently, automatic evaluation metrics used alone are likely to represent biased or not entirely correct results for the the test set of RACE-H. Note that while the results obtained in this research concern only the test set of RACE-H, the applied methodology can be used as a blueprint for analysing other subsets of RACE, or other datasets (likely with some modifications in the latter case).

\backmatter

\bmhead{Acknowledgments}
We gratefully acknowledge the Knut and Alice Wallenberg Foundation and Digital Futures (project ``SWE-QUEST'') for their generous financial support which allowed us to complete this research.

\section*{Declarations}
\begin{itemize}
\item Funding: Knut and Alice Wallenberg Foundation, Digital Futures (project ``SWE-QUEST'').
\item Competing interests: The authors declare none
\item Ethics approval: Not applicable
\item Consent to participate: Not applicable
\item Consent for publication: Not applicable
\item Availability of data and materials: RACE is a publicly available dataset and our additional annotations are available at the associated GitHub repository.
\item Code availability: the code for analysis is available at the associated GitHub repository (\url{https://github.com/dkalpakchi/EMBRACE}).
\item Authors' contributions: All authors contributed to the study conception and design. Modifications to the data evaluation model initially selected and annotation of the RACE corpus were performed by \textit{Mariia Zyrianova}.  The annotation tool (i.e. Textinator) was set up by \textit{Dmytro Kalpakchi}. Data analysis was designed by \textit{Mariia Zyrianova} and \textit{Dmytro Kalpakchi}, with the solicited code written by \textit{Dmytro Kalpakchi}. \textit{Dmytro Kalpakchi} also produced all figures included to the article. The first draft of the manuscript was written by \textit{Mariia Zyrianova} and all authors commented on previous versions of the manuscript. \textit{Johan Boye} was providing guidance during the project and revised the article critically. 

All authors reviewed the results and approved the final version of the article.

\end{itemize}

\begin{appendices}

\section{Scoring scale applied for the RACE corpus evaluation}\label{app:A}
\textbf{Final score} = TOI + TOM + NPhr + NI + NIt + NPar + IC + POD + TOC

\begin{figure}[h!]
    \centering
    \begin{tcolorbox}
    \begin{tabular}{p{10.5cm}|p{0.5cm}}
        \textbf{Category for} {\it Type of calculation (TOC)} & \textbf{Pts} \\
    \hline
    no calculations & 0 \\
    \hline
    single addition (``+'') & 1 \\
    \hline
    single subtraction (``-'') & 2 \\
    \hline
    single multiplication (``*'') & 3 \\
    \hline
    single division (``/'') & 4 \\
    \hline
    multiple operations & 5
    \end{tabular}
    \end{tcolorbox}
    \label{fig:toc}
\end{figure}

\begin{figure}[h!]
    \centering
    \begin{tcolorbox}
    \begin{tabular}{p{10.5cm}|p{0.5cm}}
        \textbf{Categories for} {\it Type of information (TOI)} & \textbf{Pts} \\
    \hline
     person, animal, place, group, thing & 1 \\
     \hline
     amount, time, attribute, action, location, type/kind, procedure, part & 2 \\
     \hline
     manner, goal, purpose, condition, predicate adjective, function, alternative, attempt, sequence, pronominal reference, verification, assertion, problem, solution, role, process  & 3 \\
     \hline
     cause, reason, result, effect, justification, evidence, similarity, opinion, explanation, theme, pattern & 4 \\
     \hline
     equivalent, difference, definition, advantage; indeterminate & 5 \\
    \end{tabular}
    \end{tcolorbox}
    \label{fig:toi}
\end{figure}

\begin{figure}[h!]
    \centering
    \begin{tcolorbox}
    \begin{tabular}{p{10.5cm}|p{0.5cm}}
        \textbf{Categories for {\it{Type of match (TOM)}}} & \textbf{Pts} \\
    \hline
     the relations between the question and text and between the text and the answer are \textit{both literal} & 0.5 \\
     \hline
     \textit{either} the relation between the question and text or between the text and the answer \textit{is literal}, while \textit{the other is synonymous} & 1 \\
     \hline 
     the relations between the question and text and between the text and the answer are \textit{both synonymous} & 1.5 \\
     \hline
     the relation between the question and text or between the text and the answer requires \textit{a low-level text-based inference (LLTI)} while \textit{the other requires literal match} & 2 \\
     \hline
     the relation between the question and text or between the text and the answer requires \textit{a low-level text-based inference (LLTI)} while \textit{the other requires synonymous match} & 2.5 \\
     \hline
     the relations between the question and text and between the text and the answer \textit{both require a low-level text-based inference (LLTI)} & 3 \\
     \hline
     \textit{either} the relation between the question and the text or between the text and the answer \textit{requires a high-level (HLTI) text-based inference} & 4 \\
     \hline      
     the relation between the text, the question, and the answer requires the reader to \textit{generate the appropriate interpretive framework} to relate the three & 5 \\
    \end{tabular}
    \end{tcolorbox}
    \label{fig:tom}
\end{figure}

\begin{figure}[h!]
\begin{subfigure}[h]{0.5\textwidth}
    \centering
    \begin{tcolorbox}
    
    {\it Number of items transparency (NIt)}\\
    
    \begin{tabular}{p{4cm}|p{0.5cm}}
        \textbf{Category} & \textbf{Pts} \\
    \hline
     specified & 0 \\
     \hline
     unspecified & 1
    \end{tabular}
    \end{tcolorbox}
    \label{fig:nit}
\end{subfigure}
\begin{subfigure}[h]{0.5\textwidth}
    \centering
    \begin{tcolorbox}
    
    {\it Number of req. paragraphs (NPar)}\\
    
    \begin{tabular}{p{4cm}|p{0.5cm}}
        \textbf{Category} & \textbf{Pts} \\
    \hline
     within paragraph (1)& 0 \\
     \hline
     between paragraphs (2+) & 1 \\
    \end{tabular}
    \end{tcolorbox}
    \label{fig:np}
\end{subfigure}
\end{figure}

\begin{figure}[h!]
\begin{subfigure}[h]{0.5\textwidth}
    \centering
    \begin{tcolorbox}
    
    {\it Number of items (NI)}\\\\
    \begin{tabular}{p{3.75cm}|p{0.5cm}}
        \textbf{Category} & \textbf{Pts} \\
    \hline
     one (``1'') & 0 \\
     \hline
     two (``2'') & 1 \\
     \hline
     three-four (``3-4'') & 2 \\
     \hline
     five or more (``5+'') & 3 \\
    \end{tabular}
    \end{tcolorbox}
    \label{fig:ni}
\end{subfigure}
\begin{subfigure}[h]{0.5\textwidth}
    \centering
    \begin{tcolorbox}
    
    {\it Number of phrases (NPhr)}\\\\
    \begin{tabular}{p{4.4cm}|p{0.5cm}}
        \textbf{Category} & \textbf{Pts} \\
    \hline
     one (``1'') & 0 \\
     \hline
     two (``2'') & 1 \\
     \hline
     three (``3'') & 2 \\
     \hline
     four or more (``4+'') & 3 \\
    \end{tabular}
    \end{tcolorbox}
    \label{fig:nphr}
\end{subfigure}
\end{figure}

\begin{figure}[h!]
    \centering
    \begin{tcolorbox}
    \begin{tabular}{p{10.5cm}|p{0.5cm}}
        \textbf{Category for} {\it{Plausibility of distractors (POD)}} & \textbf{Pts} \\
    \hline
     there is no distracting information in the text & 1 \\
     \hline
     distractors contain information that corresponds \textit{literally} to information in the text but \textit{not in the same paragraph as the correct answer} & 1.5 \\
     \hline
     distractors contain information that is \textit{synonymous} to information in the text but \textit{not in the same paragraph as the correct answer} & 2 \\
     \hline
     distractors contain information that represent plausible \textit{invited inferences} \textit{not based on} information related to \textit{the paragraph in which the correct answer occurs} & 3 \\
     \hline
     \textit{one distractor} in the choices contains information that is \textit{related to the information in the same paragraph as the answer} & 4 \\
     \hline
     \textit{two or more distractors} in the choices contain information that is \textit{related to the information in the same paragraph as the answer} & 5 \\
     \hline
     \textit{one or more distractors} represent plausible inferences \textit{based on information outside the text} & 5
    \end{tabular}
    \end{tcolorbox}
    \label{fig:pod}
\end{figure}
\begin{figure}[h!]
    \centering
    \begin{tcolorbox}
    \begin{tabular}{p{10.5cm}|p{0.5cm}}
        \textbf{Category for} {\it Infer condition (IC)} & \textbf{Pts} \\
    \hline
     compare or based on synthesis of features throughout paragraph & 0 \\
     \hline
     contrast or based on synthesis of features throughout paragraphs & 1
    \end{tabular}
    \end{tcolorbox}
    \label{fig:ic}
\end{figure}

\section{Glossary of errors}\label{app:B}
More examples and explanations are available in \cite[Section 3]{zyrianovam}.\\

\textbf{Additional notes} – an MCQ element contains special characters, symbols, or remarks (e.g., \textit{prefix = st1 /, s6t----}, etc.) which are not related to the main content, are unnecessary for a reader in the context of a reading comprehension test (e.g., \textit{number of words, source references}, etc.), or provide vague specifications.\\

%\textbf{Note:} explanations of the vocabulary in the texts and/or synonyms used to clarify meaning of a word/phrase are not counted as A.n..\\

\textbf{Ambiguously formulated} – the question is formulated in a way that does not make the query clear or allows two possible interpretations of the question.

Sometimes, the error can be a result of wrongly used pronouns in the question.\\

\textbf{Answerable without reading} – it is possible to answer the question by analysing alternatives, without reading the given passage, or even without reading the alternatives (often applies to questions which imply stable facts, common truth, meanings of expressions, and definitions).\\

\textbf{Extra gaps} – a piece of information is or might be missed in an MCQ element, though has no significant influence on general understanding of the content and answering the given question (if the E.g. is in the text, it might appear misleading for another question to the same text). Some of the gaps are marked with an underscore (\_).\\

\textbf{Extra spaces (punctuation)} – one or more spaces are inserted before/after a punctuation mark where they considered erroneous according to the English language rules.\\

\textbf{Extra spaces (within a word or a digit)} – one or more spaces are inserted in the middle of a word or a multi-digit number, which leads to a (misleading) spelling error or confusion in the data interpretation:

Here, web addresses are counted as one word, currency symbol and figure are counted as one digit; stable phrases and abbreviations with no fixed written form (e.g., \textit{a.m. / am / AM,} etc.) are only checked for consistent formatting within the text.\\

\textbf{Formatting inconsistency} – \textbf{\textit{for questions}}, means that the part claimed to be formatted in a certain way (e.g., underlined) cannot be found as such or differs in actual formatting;

\textbf{\textit{for alternatives}}, formatting rules (e.g., capitalisation of the first word, punctuation in the end of an alternative, etc.) differ within one and the same MCQ unit;

\textbf{\textit{for texts}}, means that formatting rules for abbreviations, numbers, or lists are not completely followed (similarly to the way the definition applies for alternatives).

The issue is treated as problematic as it might be used as a meta-clue for a correct answer (especially, when only one of the alternatives differ in formatting from the other) or cause certain confusion for a reader when interpreting the information given in the passage.\\

\textbf{Grammatical errors} – a sentence contains a faulty, uncommon, or controversial usage of the English language (e.g., pronoun disagreement, misuse of articles, grammar tenses, modifiers, etc.), or breaks the English grammar rules.\\

\textbf{Incomplete alternative(s)} – one or more of the alternatives fails to provide a comprehensible piece of information to be related to the given question and/or text.\\

\textbf{Incomplete question} – the question fails to provide a comprehensible query and/or cannot be related to the text.\\

\textbf{Incomplete text} – the text does not provide enough information to relate the question.\\

\textbf{Inconsistency between A} – alternatives within one and the same MCQ unit do not correspond in represented type of content.

\textbf{Inconsistency between Q and A} – question and one or more of the alternatives within one and the same MCQ unit represent different types of information.

\textbf{Inconsistency between T and A} – the data provided in the text do not correspond semantically to that in one or more of the alternatives.

\textbf{Inconsistency between T and Q} – the data requested by the question is not provided or cannot be inferred from the text.\\

\textbf{Misleading gap(s)} – a piece of information is missed in an MCQ element and prevents a reader from understanding of the content and/or answering the given question (if a M.g. is in the text, it often appears in the area identified as a key paragraph and a basis for one or more of the alternatives). The gaps might or might not be marked with an underscore (\_).

\textbf{Misleading spaces} – one or more spaces are inserted before/after a punctuation mark where they considered erroneous according to the English language rules, and lead to (misleading) spelling errors.\\

\textbf{Misleading spelling errors} – a word/phrase is formed incorrectly in terms of choice of letters, order of letters, or usage of special characters/symbols (e.g., hyphens, apostrophes), which makes it impossible to recognise the affected word/phrase properly or results in a different word/phrase implying other meaning(s).\\

\textbf{Missing spaces (between words and/or digits)} – lack of a blank area to separate two or more words or a word and a digit; can also lead to (misleading) spelling errors.\\

\textbf{Missing spaces (punctuation)} – lack of a blank area to separate one or more punctuation mark from a following word, digit, or another punctuation mark, according to the English language punctuation rules.\\

\textbf{OCR errors} – faults caused by inaccurate optical character recognition process; often leads to (misleading) spelling errors, digits inserted in a word, and letters inserted in a multi-digit number. \\

\textbf{Overlapping alternatives} – more than one alternative satisfy the conditions stated in the question and thus result in more than one correct answer to one and the same question.\\

\textbf{Punctuation errors} – wrong choice or usage of punctuation symbols, according to the English language punctuation rules (e.g., periods used instead of commas and vice versa, unnecessary or missed punctuation marks in a sentence). \\

\textbf{Spelling errors (except hyphens and contractions)} – a word/phrase is formed incorrectly in terms of choice of letters, order of letters, capitalisation or usage of special characters/symbols according to the rules of British or American English, depending on the language of each given text.\\ 

\textbf{Spelling errors (hyphens, contractions)} – incorrect usage of hyphens and contractions (or missing symbols) which is often a result of extra or missing spaces and might lead to misleading spelling errors.\\

\textbf{Subjective formulation} – the question or one or more of the alternatives requires evaluation of an object or phenomenon basing on a reader’s personal opinion, feelings or experience, where the answer will likely lack grounds for support or declination, and consequently, for objective evaluation.\\

\textbf{Syntax errors} – incorrect word order in one or more sentences, or unnecessary repetition of one or more syntactic unit.\\

\textbf{Time-dependent} – the requested information is related to the period of time that passed from the date of publication of the text or from when an event discussed in the passage happened till the text has been published; this makes the correct answer not stable and requires changes in corresponding periods of time (e.g., months, years, etc.).



%\section{Example of another appendix section}\label{secA2}%
 %Appendices may be used for helpful, supporting or essential material that would otherwise 
 %clutter, break up or be distracting to the text. Appendices %can consist of sections, figures, 
 %tables and equations etc.

\end{appendices}


\bibliography{sn-bibliography}% common bib file

\end{document}